# CIXL2: A Crossover Operator for Evolutionary Algorithms Based on Population Features

**Domingo Ortiz-Boyer** DORTIZ@UCO.ES
**César Hervás-Martínez** CHERVAS@UCO.ES
**Nicolás García-Pedrajas** NPEDRAJAS@UCO.ES
*Department of Computing and Numerical Analysis*
*University of Córdoba, Spain*

## Abstract

In this paper we propose a crossover operator for evolutionary algorithms with real values that is based on the statistical theory of population distributions. The operator is based on the theoretical distribution of the values of the genes of the best individuals in the population. The proposed operator takes into account the localization and dispersion features of the best individuals of the population with the objective that these features would be inherited by the offspring. Our aim is the optimization of the balance between exploration and exploitation in the search process.

In order to test the efficiency and robustness of this crossover, we have used a set of functions to be optimized with regard to different criteria, such as, multimodality, separability, regularity and epistasis. With this set of functions we can extract conclusions in function of the problem at hand. We analyze the results using ANOVA and multiple comparison statistical tests.

As an example of how our crossover can be used to solve artificial intelligence problems, we have applied the proposed model to the problem of obtaining the weight of each network in a ensemble of neural networks. The results obtained are above the performance of standard methods.

## 1. Introduction

Evolutionary algorithms (EAs) are general purpose searching methods. The selection process and the crossover and mutation operators establish a balance between the exploration and exploitation of the search space which is very adequate for a wide variety of problems whose solution presents difficulties that are insolvable using classical methods. Most of these problems are defined in continuous domains, so the evolutionary algorithms applied use real values, namely, evolution strategies (EPs), real-coded genetic algorithms (RCGAs), and evolutionary programming (EP). For these paradigms the precision of the solution does not depend on the coding system, as in binary coded genetic algorithms, but on the precision of the computer system where the algorithms are run.

The selection process drives the searching towards the regions of the best individuals. The mutation operator randomly modifies, with a given probability, one or more genes of a chromosome, thus increasing the structural diversity of the population. As we can see, it is clearly an exploration operator, that helps to recover the genetic diversity lost during the selection phase and to explore new solutions avoiding premature convergence. In this way, the probability of reaching a given point in the search space is never zero. This operator, in fact, implements a random search whose well-studied features are useful in the field of evolutionary computation.

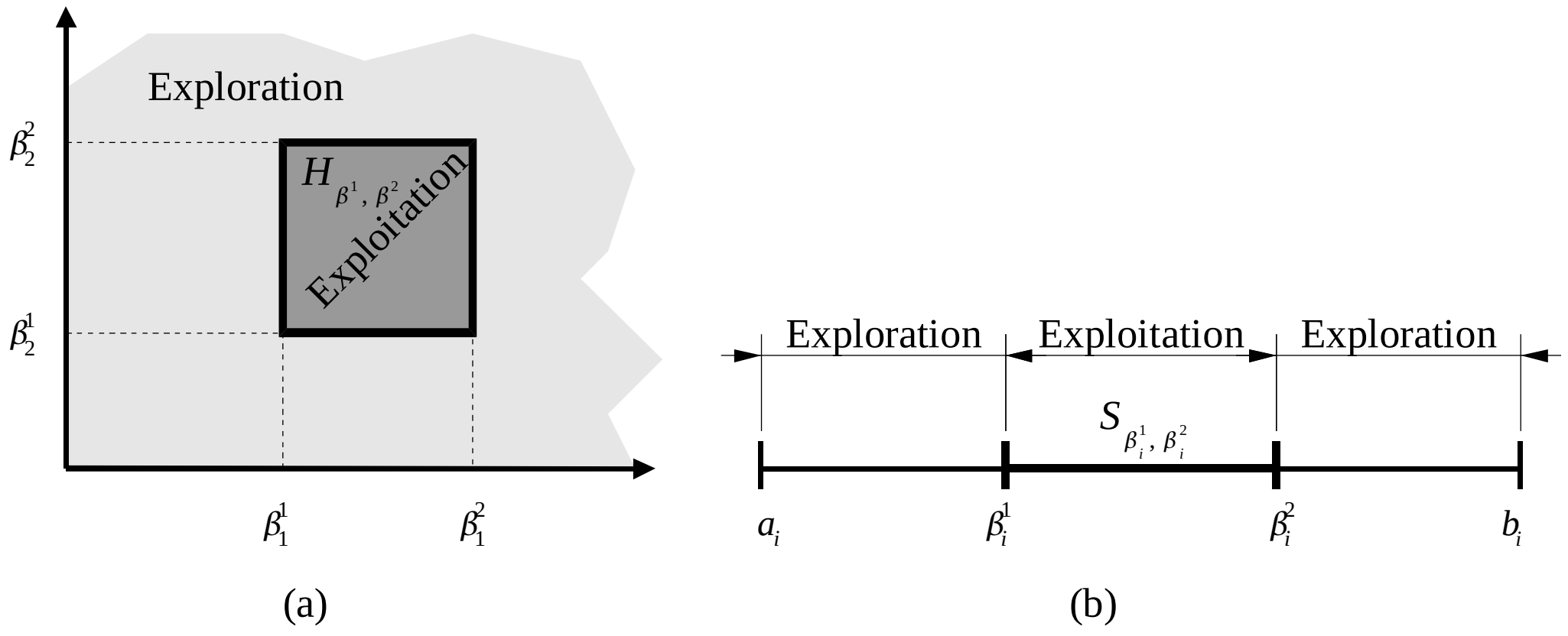

Figure 1: (a) Hypercube defined by the first two genes of the parents; (b) Representation of the segment defined by the $i$th genes of two chromosomes.

The crossover operator combines the genes of two or more parents to generate better offspring. It is based on the idea that the exchange of information between good chromosomes will generate even better offspring. The effect of the crossover operator can be studied from two different points of view: at chromosome level and at gene level. The effect of the crossover operator at chromosome level can be considered in a geometric way. Given two parents $\beta^1 = \{\beta_1^1, \beta_2^1\}$ and $\beta^2 = \{\beta_1^2, \beta_2^2\}$ with two genes, we denote by $H_{\beta^1\beta^2}$ the hypercube defined by their genes (Figure 1a). At gene level the representation would be linear, defining in this case a segment or interval $S_{\beta_i^1,\beta_i^2}$ for each pair of genes (Figure 1b). Most crossover operators generate individuals in the exploitation zones, $S_{\beta_i^1,\beta_i^2}$ or $H_{\beta^1\beta^2}$. In this way, the crossover operator implements a depth search or exploitation, leaving the breadth search or exploration for the mutation operator.

This policy, intuitively very natural, makes the population converge to values within the hypercubes defined by their parents, producing a rapid decrease in the population diversity which could end up in a premature convergence to a non-optimal solution. Recent studies on BLX-$\alpha$ crossover (Eshelman & Schaffer, 1993), the crossover based on fuzzy connectives (Herrera, Herrera-Viedma, Lozano, & Verdegay, 1994), and fuzzy recombination (Voigt, Mühlenbein, & Cvetkovic, 1995), have confirmed the good performance of those crossover operators that also generate individuals in the exploration zone. These operators avoid the loss of diversity and the premature convergence to inner points of the search space, but also the generation of new individuals in the exploration zone could slow the search process. For this reason, the crossover operator should establish an adequate balance between exploration (or interpolation) and exploitation (or extrapolation), and generate offspring in the exploration and exploitation zones in the correct proportion.

Establishing a balance between exploration and exploitation is important, but it is also important that such a balance is self-adaptive (Kita, 2001; Beyer & Deb, 2001; Deb & Beyer, 2001), that is, it must guarantee that the dispersion of the offspring depends on the dispersion of the parents. So, two close parents must generate close offspring, and two distant parents must generate distant offspring. The control of dispersion in the crossover based on fuzzy connectives is based on the generation of offspring using the fuzzy connectives $t$-norms, $t$-conorms, average functions, and a generalized operator of compensation (Mizumoto, 1989). In fuzzy recombination the offspring is generated using two triangular distributions whose averages derive from each of the genes of the two parents. In BLX-$\alpha$ we have the same probability of generating an offspring between the parents, and in an area close to the parents whose amplitude is modulated by the $\alpha$ parameter.

Ono and Kobayashi (1997) have proposed a Unimodal Normally Distributed Crossover (UNDX), where three parents are used to generate two or more children. The children are obtained using an ellipsoidal distribution where one axis is the segment that joins the two parents and the extent of the orthogonal direction is decided by the perpendicular distance of the third parent from the axis. The authors claim that this operator should preserve the statistics of the population. This crossover is also self-adaptive, but it differs from BLX-$\alpha$ by the fact that it is more probable to generate offspring near the average of the first two parents.

Another self-adaptive crossover is the Simulated Binary Crossover (SBX) (Deb & Agrawal, 1995). Based on the search features of the single-point crossover used in binary-coded genetic algorithms, this operator respects the interval schemata processing, in the sense that common interval schemata of the parents are preserved in the offspring. The SBX crossover puts the stress on generating offspring near the parents. So, the crossover guarantees that the extent of the children is proportional to the extent of the parents, and also favors that near parent individuals are monotonically more likely to be chosen as children than individuals distant from the parents.

The main goal of this paper is to propose a crossover operator that avoids the loss of diversity of the population of individuals, and, at the same time, favors the speed of convergence of the algorithm. These two goals are, at first, conflicting; their adequate balance is controlled by two of the basic features of the crossover operator: i) the balance between exploration and exploitation and, ii) the self-adaptive component. These two features make the evolutionary algorithms avoid premature convergence and favor local fine-tuning. Both attributes are highly appreciated in any search algorithm.

In most current crossover operators, the features of the offspring depend on the features of just a few parents. These crossovers do not take into account population features such as localization and dispersion of the individuals. The use of these statistical features of the population may help the convergence of the population towards the global optimum.

The crossover operator implements basically a depth or exploitative search, just like other methods such as steepest gradient descent, local search or simulated annealing, but in these three search methods the algorithm takes the quality of the solutions into account. So, it is reasonable to think that it is also convenient for the crossover operator to consider the performance on the individuals involved in the crossover operation. This idea is already implemented by some heuristic crossovers (Wright, 1991).

Nevertheless, following the previous line of argument, it seems rather poor to use just two parents, and not to consider the most promising directions towards which it would be advisable to drive the search. That is, instead of using a local heuristic that uses two individuals, involving the whole population or an adequate subset in the determination of the direction of the search whose features would be specially suitable.

Motivated by this line of argument, in this paper we propose a crossover operator, which will be called *Confidence Interval Based Crossover using $L_2$ Norm* (CIXL2). On the one hand, it takes advantage of the selective component that is derived from the extraction of the features of the best $n$ individuals of the population and that indicates the direction of the search, and on the other hand, it makes a self-adaptive sampling around those features whose width depends on the number of best individuals, dispersion of those best individuals, confidence coefficient, and localization of the individuals that participate in the crossover. Now, the exploitation region is not the area between the two parents that are involved in the crossover, but the area defined by the confidence interval built from the $n$ best individuals of the population; and the exploratory region is the rest of the search domain. To the previous concepts of exploration and exploitation, merely geometrical, is added a probabilistic component that depends on the population features of the best individuals.

Estimation of Distribution Algorithms (EDAs) or Probabilistic Model-Building Evolutionary Algorithms (Mühlenbein & Paa$\beta$, 1998; Mühlenbein, Mahnig, & Rodriguez, 1999) are based on a, seemingly, similar idea. These algorithms do not have mutation and crossover operators. After every generation the population distribution of the selected individuals is estimated and the new individuals are obtained sampling this estimated distribution. However, the underlying idea behind our crossover is the extraction of population features, mean and standard deviation, in order to detect the regions where there is a higher probability of getting the best individuals. In order to perform the crossover, we create three virtual parents that represent the localization estimator mean, and the bounds of the confidence interval from which, with a certain confidence degree, this localization estimator takes the values. In this way, the children generated from these three parents will inherit the features of the best individuals of the population.

The rest of the paper is organized as follows: Section 2 explains the definition of CIXL2 and its features; Section 3 discusses the problem of the selection of the test sets, and justifies the use of a test set based on the one proposed by Eiben and Bäck (1997a); Section 4 describes the experimental setup of evolutionary algorithm (RCGA) used in the tests; Section 5 studies the optimal values of the parameters of CIXL2; Section 6 compares the performance of CIXL2 against other crossovers; Section 7 compares CIXL2 with EDAs; Section 8 describes the application of RCGAs with CIXL2 to neural network ensembles; and, finally, Section 9 states the conclusions of our paper and the future research lines.

## 2. CIXL2 Operator

In this section we will explain the theoretical base that supports the defined crossover operator, and then we will define the crossover. We will use an example to explain the dynamics of a population subject to this crossover operator.

### 2.1 Theoretical Foundation

In this section we will study the distribution of the $i$-th gene and the construction of a confidence interval for to the localization parameter associated with that distribution.

Let $\boldsymbol{\beta}$ be the set of $N$ individuals with $p$ genes that make up the population and $\boldsymbol{\beta^*} \subset \boldsymbol{\beta}$ the set of the best $n$ individuals. If we assume that the genes $\beta_i^*$ of the individuals belonging to $\boldsymbol{\beta^*}$ are independent random variables with a continuous distribution $H(\beta_i^*)$ with a localization parameter $\mu_{\beta_i^*}$, we can define the model

$$\beta_i^* = \mu_{\beta_i^*} + e_i, \quad \text{for} \;\; i = 1, ..., p, \tag{1}$$

being $e_i$ a random variable. If we suppose that, for each gene $i$, the best $n$ individuals form a random sample $\{\beta_{i,1}^*, \beta_{i,2}^*, ..., \beta_{i,n}^*\}$ of the distribution of $\beta_i^*$, then the model takes the form

$$\beta_{ij}^* = \mu_{\beta_i^*} + e_{ij}, \quad \text{for} \;\; i = 1, ..., p \;\; \text{and} \;\; j = 1, ..., n. \tag{2}$$

Using this model, we analyze an estimator of the localization parameter for the $i$-th gene based on the minimization of the dispersion function induced by the $L_2$ norm. The $L_2$ norm is defined as

$$\|e_i\|_2^2 = \sum_{j=1}^{n} (e_{ij})^2, \tag{3}$$

hence the associated dispersion induced by the $L_2$ norm in the model 2 is

$$D_2(\mu_{\beta_i^*}) = \sum_{j=1}^{n} (\beta_{ij}^* - \mu_{\beta_i^*})^2, \tag{4}$$

and the estimator of the localization parameter $\mu_{\beta_i^*}$ is:

$$\hat{\mu}_{\beta_i^*} = \arg\min D_2(\mu_{\beta_i^*}) = \arg\min \sum_{j=1}^{n} (\beta_{ij}^* - \mu_{\beta_i^*})^2. \tag{5}$$

Using for minimization the steepest gradient descent method,

$$S_2(\mu_{\beta_i^*}) = -\frac{\partial D_2(\mu_{\beta_i^*})}{\partial \mu_{\beta_i^*}}, \tag{6}$$

we obtain

$$S_2(\mu_{\beta_i^*}) = 2 \sum_{j=1}^{n} (\beta_{ij}^* - \mu_{\beta_i^*}), \tag{7}$$

and making (7) equal to 0 yields

$$\hat{\mu}_{\beta_i^*} = \frac{\sum_{j=1}^{n} \beta_{ij}^*}{n} = \bar{\beta}_i^*. \tag{8}$$

So, the estimator of the localization parameter for the $i$-th gene based on the minimization of the dispersion function induced by the $L_2$ norm is the mean of the distribution of $\beta_i^*$ (Kendall & Stuart, 1977), that is, $\hat{\mu}_{\beta_i^*} = \bar{\beta}_i^*$.

The sample mean estimator is a linear estimator[1], so it has the properties of unbiasedness[2] and consistency[3], and it follows a normal distribution $N(\mu_{\beta_i^*}, \sigma^2_{\beta_i^*}/n)$ when the distribution of the genes $H(\beta_i^*)$ is normal. Under this hypothesis, we construct a bilateral confidence interval for the localization of the genes of the best $n$ individuals, using the studentization method, the mean as the localization parameter,and the standard deviation $S_{\beta_i^*}$ as the dispersion parameter:

$$I^{CI} = \left[\bar{\beta}_i^* - t_{n-1,\alpha/2}\frac{S_{\beta_i^*}}{\sqrt{n}}; \bar{\beta}_i^* + t_{n-1,\alpha/2}\frac{S_{\beta_i^*}}{\sqrt{n}}\right] \tag{9}$$

where $t_{n-1,\alpha/2}$ is the value of Student's $t$ distribution with $n-1$ degrees of freedom, and $1-\alpha$ is the confidence coefficient, that is, the probability that the interval contains the true value of the population mean.

### 2.2 CIXL2 Definition

From this definition of the confidence interval, we define three intervals to create three "virtual" parents, formed by the lower limits of the confidence interval of each gene, $CILL$[4], the upper limits, $CIUL$[5], and the means $CIM$[6]. These parents have the statistical information of the localization features and dispersion of the best individuals of the population, that is, the genetic information the fittest individuals share. Their definition is:

$$\begin{aligned} CILL &= (CILL_1, \dots, CILL_i, \dots CILL_p) \\ CIUL &= (CIUL_1, \dots, CIUL_i, \dots CIUL_p) \\ CIM &= (CIM_1, \dots, CIM_i, \dots CIM_p), \end{aligned} \tag{10}$$

where

$$\begin{aligned} CILL_i &= \bar{\beta}_i^* - t_{n-1,\alpha/2}\frac{S_{\beta_i^*}}{\sqrt{n}} \\ CIUL_i &= \bar{\beta}_i^* + t_{n-1,\alpha/2}\frac{S_{\beta_i^*}}{\sqrt{n}} \\ CIM_i &= \bar{\beta}_i. \end{aligned} \tag{11}$$

The $CILL$ and $CIUL$ individuals divide the domain of each gene into three subintervals: $D_i \equiv I_i^L \cup I_i^{CI} \cup I_i^U$, where $I_i^L \equiv [a_i, CILL_i)$; $I_i^{CI} \equiv [CILL_i, CIUL_i]$; $I_i^U \equiv (CIUL_i, b_i]$; being $a_i$ and $b_i$ the bounds of the domain (see Figure 2).

The crossover operator creates one offspring $\beta^s$, from an individual of the population $\beta^f \in \boldsymbol{\beta}$, randomly selected, and one of the individuals $CILL$, $CIUL$ or $CIM$, depending on the localization of $\beta^f$, as follows:

1. *It is a linear combination of the sample values.*
2. *An estimator $\hat{\theta}$ is an unbiased estimator of $\theta$ if the expected value of the estimator is the parameter to be estimate: $E[\hat{\theta}] = \theta$.*
3. *A consistent estimator is an estimator that converges in probability to the quantity being estimated as the sample size grows.*
4. *Confidence Interval Lower Limit.*
5. *Confidence Interval Upper Limit.*
6. *Confidence Interval Mean.*

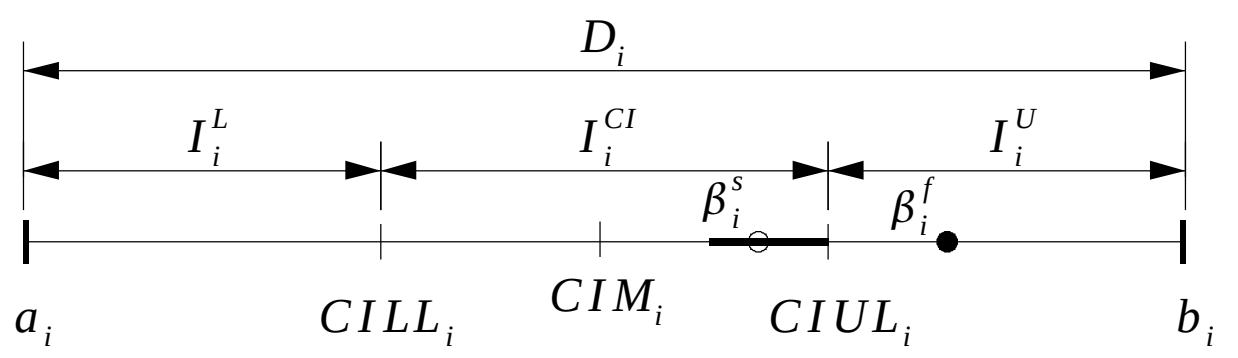

Figure 2: An example of confidence interval based crossover

- $\beta_i^f \in I_i^L$: if the fitness of $\beta^f$ is higher than CILL, then $\beta_i^s = r(\beta_i^f - CILL_i) + \beta_i^f$, else $\beta_i^s = r(CILL_i - \beta_i^f) + CILL_i$.

- $\beta_i^f \in I_i^{CI}$: if the fitness of $\beta^f$ is higher than CIM, then $\beta_i^s = r(\beta_i^f - CIM_i) + \beta_i^f$, else $\beta_i^s = r(CIM_i - \beta_i^f) + CIM_i$.

- $\beta_i^f \in I_i^U$: if the fitness of $\beta^f$ is higher than CIUL, then $\beta_i^s = r(\beta_i^f - CIUL_i) + \beta_i^f$, else $\beta_i^s = r(CIUL_i - \beta_i^f) + CIUL_i$ (this case can be seen in Figure 2).

where $r$ is a random number in the interval $[0, 1]$.

With this definition, the offspring always takes values in the direction of the best of the two parents but never between them. If the virtual individual is one of the bounds of the confidence interval and is better than the other parent, the offspring is generated in the direction of the confidence interval where it is more likely to generate better individuals. If the virtual individual is worse than the other parent, the offspring is generated near the other parent in the opposite direction of the confidence interval. On the other hand, if a parent selected from the population is within the confidence interval, the offspring can be outside the interval – but always in its neighborhood – if the fitness of the center of the confidence interval is worse. This formulation tries to avoid a shifting of the population towards the confidence interval, unless this shifting means a real improvement of the fitness in the population.

If $\beta^f$ is distant from the other parent, the offspring will probably undergo a marked change, and if both parents are close, the change will be small. The first circumstance will be likely to occur in the first stages of the evolutionary process, and the second one in the final stages.

The width of the interval $I^{CI}$ depends on the confidence coefficient, $1 - \alpha$, the number of best individuals, $n$, and the dispersion of the best individuals. In the first stages of the evolution, the dispersion will be large, specially for multimodal functions, and will decrease together with the convergence of the genetic algorithm. These features allow the balance between exploitation and exploration to adjust itself dynamically. The crossover will be more exploratory at the beginning of the evolution, avoiding a premature convergence, and more exploitative at the end, allowing a fine tuning. The parameters $n$ and $1-\alpha$ regulate the dynamics of the balance favoring a higher or lower degree of exploitation. That suggests the CIXL2 establishes a self-adaptive equilibrium between exploration and exploitation based on the features that share, with a certain confidence degree $1 - \alpha$, the best $n$ individuals of the population. A preliminary theoretical study of this aspect is carried out by Hervás-Martínez and Ortiz-Boyer (2005).

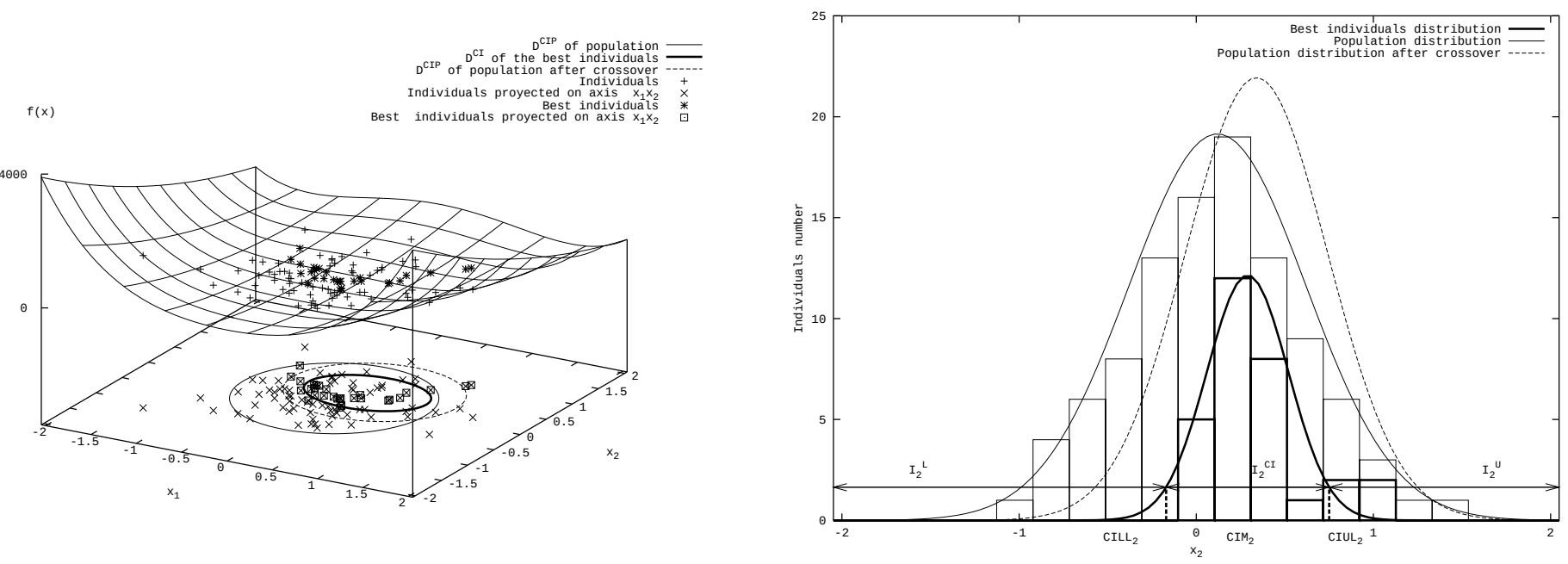

(a) (b)

Figure 3: Effect of the CIXL2 crossover over a population used for the minimization of the Rosenbrock function with two variables

### 2.3 Crossover Dynamics

Figure 3 shows a simulation of the behavior of the crossover for the optimization of Rosenbrock function (Eiben & Bäck, 1997b) with two variables. On Figure 3a, we observe how most of the individuals are within the domain $D^{CIP}$; while the best $n$ are within the confidence domain $D^{CI} \equiv I_1^{CI} \times I_2^{CI}$. $D^{CI}$ is shifted towards the minimum of the function placed in $(1, 1)$, the domain $D^{CIP}$ of the new population, generated after applying CIXL2, will be shifted to the optimum. This displacement will be higher in the first stages of evolution, and will decrease during evolution. It may be modulated by the parameters $n$ and $1 - \alpha$.

Figure 3a shows how the population, after applying the crossover operator, is distributed in a region nearer the optimum whose diversity depends on the parameters of the operator.

Figure 3b shows how the whole population and the $n$ best individuals are distributed. As we can see, the distribution of the best $n$ individuals keeps the features of the distribution of the population, but it is shifted to the optimum. The shifting towards the optimum will be more marked if the value of $n$ is small. The tails of the distribution of the best individuals will be larger if the dispersion of the best individuals is also large, and smaller if they are concentrated in a narrow region. The size of these tails also depends on the features of the problem, the stage of the evolution, and the particular gene considered. The effect of the crossover on the distribution of the population is to shift the distribution towards the best $n$ individuals and to stretch the distribution modulately depending on the amplitude of the confidence interval. The parameters $n$ and $1 - \alpha$ are responsible for the displacement and the stretching of the region where the new individuals will be generated.

If $n$ is small, the population will move to the most promising individuals quickly. This may be convenient for increasing the convergence speed in unimodal functions. Nevertheless, it can produce a premature convergence to suboptimal values in multimodal functions. If $n$ is large, both the shifting and the speed of convergence will be smaller. However, the evolutionary process will be more robust, this feature being perfectly adequate for the optimization of multimodal, non-separable, highly epistatic functions.

The parameter $n$ is responsible for the selectiveness of the crossover, as it determines the region where the search will be directed. The selection is regulated by the parameter $1 - \alpha$. This parameter bounds the error margin of the crossover operator in order to obtain a search direction from the feature that shares the best individuals of the population.

## 3. Benchmark Problems

In the field of evolutionary computation, it is common to compare different algorithms using a large test set, especially when the test involves function optimization (Gordon & Whitley, 1993). However, the effectiveness of an algorithm against another algorithm cannot be measured by the number of problems that it solves better. The *"no free lunch"* theorem (Wolpert & Macready, 1995) shows that, if we compare two searching algorithms with all possible functions, the performance of any two algorithms will be , on average, the same . As a result, attempting to design a perfect test set where all the functions are present in order to determine whether an algorithm is better than another for every function, is a fruitless task.

That is the reason why, when an algorithm is evaluated, we must look for the kind of problems where its performance is good, in order to characterize the type of problems for which the algorithm is suitable. In this way, we have made a previous study of the functions to be optimized for constructing a test set with fewer functions and a better selection (Whitley, Mathias, Rana, & Dzubera, 1995; Salomon, 1996). This allows us to obtain conclusions of the performance of the algorithm depending on the type of function.

Taking into account this reasoning, the test set designed by Eiben and Bäck (1997b) is very adequate. The test set has several well characterized functions that will allow us to obtain and generalize, as far as possible, the results regarding the kind of function involved. Nevertheless, we have added two functions to the test set with the aim of balancing the number of functions of each kind. These two new functions are the function of Rosenbrock (Rosenbrock, 1960) extended to $p$ dimensions and the function of Schwefel (Schwefel, 1981); both of them have been widely used in evolutive optimization literature. Table 1 shows the expression of each function and a summary of its features: separability, multimodality, and regularity.

A function is multimodal if it has two or more local optima. A function of $p$ variables is separable if it can be rewritten as a sum of $p$ functions of just one variable (Hadley, 1964). The separability is closely related to the concept of epistasis or interrelation among the variables of the function. In the field of evolutionary computation, the epistasis measures how much the contribution of a gene to the fitness of the individual depends on the values of other genes.

Non separable functions are more difficult to optimize as the accurate search direction depends on two or more genes. On the other hand, separable functions can be optimized for each variable in turn. The problem is even more difficult if the function is also multimodal. The search process must be able to avoid the regions around local minima in order to approximate, as far as possible, the global optimum. The most complex case appears when the local optima are randomly distributed in the search space.

| Function | Definition | Multimodal? | Separable? | Regular? |
|---|---|---|---|---|
| Sphere | $f_{Sph}(\mathbf{x}) = \sum_{i=1}^{p} x_i^2$<br>$x_i \in [-5.12, 5.12]$<br>$\mathbf{x}^* = (0, 0, \ldots, 0);\ f_{Sph}(\mathbf{x}^*) = 0$ | no | yes | n/a |
| Schwefel's double sum | $f_{SchDS}(\mathbf{x}) = \sum_{i=1}^{p} \left(\sum_{j=1}^{i} x_j\right)^2$<br>$x_i \in [-65.536, 65.536]$<br>$\mathbf{x}^* = (0, 0, \ldots, 0);\ f_{SchDS}(\mathbf{x}^*) = 0$ | no | no | n/a |
| Rosenbrock | $f_{Ros}(\mathbf{x}) = \sum_{i=1}^{p-1}[100(x_{i+1} - x_i^2)^2 + (x_i - 1)^2]$<br>$x_i \in [-2.048, 2.048]$<br>$\mathbf{x}^* = (1, 1, \ldots, 1);\ f_{Ros}(\mathbf{x}^*) = 0$ | no | no | n/a |
| Rastrigin | $f_{Ras}(\mathbf{x}) = 10p + \sum_{i=1}^{p}(x_i^2 - 10\cos(2\pi x_i))$<br>$x_i \in [-5.12, 5.12]$<br>$\mathbf{x}^* = (0, 0, \ldots, 0);\ f_{Ras}(\mathbf{x}^*) = 0$ | yes | yes | n/a |
| Schwefel | $f_{Sch}(\mathbf{x}) = 418.9829 \cdot p + \sum_{i=1}^{p} x_i sin\left(\sqrt{\lvert x_i \rvert}\right)$<br>$x_i \in [-512.03, 511.97]$<br>$\mathbf{x}^* = (-420.9687, \ldots, -420.9687);\ f_{Sch}(\mathbf{x}^*) = 0$ | yes | yes | n/a |
| Ackley | $f_{Ack}(\mathbf{x}) = 20 + e - 20exp\left(-0.2\sqrt{\frac{1}{p}\sum_{i=1}^{p} x_i^2}\right) -$ $exp\left(\frac{1}{p}\sum_{i=1}^{p} cos(2\pi x_i)\right)$<br>$x_i \in [-30, 30]$<br>$\mathbf{x}^* = (0, 0, \ldots, 0);\ f_{Ack}(\mathbf{x}^*) = 0$ | yes | no | yes |
| Griewangk | $f_{Gri}(\mathbf{x}) = 1 + \sum_{i=1}^{p} \frac{x_i^2}{4000} - \prod_{i=1}^{p} cos\left(\frac{x_i}{\sqrt{i}}\right)$<br>$x_i \in [-600, 600]$<br>$\mathbf{x}^*(0, 0, \ldots, 0);\ f_{Gri}(\mathbf{x}^*) = 0$ | yes | no | yes |
| Fletcher Powell | $f_{Fle}(\mathbf{x}) = \sum_{i=1}^{p}(A_i - B_i)^2$<br>$A_i = \sum_{j=1}^{p}(a_{ij} sin\alpha_j + b_{ij} cos\alpha_j)$<br>$B_i = \sum_{j=1}^{p}(a_{ij} sinx_j + b_{ij} cosx_j)$<br>$x_i, \alpha_i \in [-\pi, \pi];\ a_{ij}, b_{ij} \in [-100, 100]$<br>$\mathbf{x}^* = \alpha;\ f_{Fle}(\mathbf{x}^*) = 0$ | yes | no | no |
| Langerman | $f_{Lan}(\mathbf{x}) = -\sum_{i=1}^{m} c_i \cdot exp\left(-\frac{1}{\pi}\sum_{j=1}^{p}(x_j - a_{ij})^2\right) \cdot$ $cos\left(\pi\sum_{j=1}^{p}(x_j - a_{ij})^2\right)$<br>$x_i \in [0, 10];\ m = p$<br>$\mathbf{x}^* = random;\ f_{Lan}(\mathbf{x}^*) = random$ | yes | no | no |

Table 1: Definition of each function together with its features

The dimensionality of the search space is another important factor in the complexity of the problem. A study of the dimensionality problem and its features was carried out by Friedman (1994). In order to establish the same degree of difficulty in all the problems, we have chosen a search space of dimensionality $p = 30$ for all the functions.

Sphere function has been used in the development of the theory of evolutionary strategies (Rechenberg, 1973), and in the evaluation of genetic algorithms as part of the test set proposed by De Jong (1975). Sphere, or De Jong's function F1, is a simple and strongly convex function. Schwefel's double sum function was proposed by Schwefel (1995). Its main difficulty is that its gradient is not oriented along their axis due to the epistasis among their variables; in this way, the algorithms that use the gradient converge very slowly. Rosenbrock function (Rosenbrock, 1960), or De Jong's function F2, is a two dimensional function with a deep valley with the shape of a parabola of the form $x_1^2 = x_2$ that leads to the global minimum. Due to the non-linearity of the valley, many algorithms converge slowly because they change the direction of the search repeatedly. The extended version of this function was proposed by Spedicato (1975). Other versions have been proposed (Oren, 1974; Dixon, 1974). It is considered by many authors as a challenge for any optimization algorithm (Schlierkamp-Voosen, 1994). Its difficulty is mainly due to the non-linear interaction among its variables.

Rastrigin function (Rastrigin, 1974) was constructed from Sphere adding a modulator term $\alpha \cdot cos(2\pi x_i)$. Its contour is made up of a large number of local minima whose value increases with the distance to the global minimum. The surface of Schwefel function (Schwefel, 1981) is composed of a great number of peaks and valleys. The function has a second best minimum far from the global minimum where many search algorithms are trapped. Moreover, the global minimum is near the bounds of the domain.

Ackley, originally proposed by Ackley (1987) and generalized by Bäck (1993), has an exponential term that covers its surface with numerous local minima. The complexity of this function is moderated. An algorithm that only uses the gradient steepest descent will be trapped in a local optima, but any search strategy that analyzes a wider region will be able to cross the valley among the optima and achieve better results. In order to obtain good results for this function, the search strategy must combine the exploratory and exploitative components efficiently. Griewangk function (Bäck, Fogel, & Michalewicz, 1997) has a product term that introduces interdependence among the variables. The aim is the failure of the techniques that optimize each variable independently. As in Ackley function, the optima of Griewangk function are regularly distributed.

The functions of Fletcher-Powell (Fletcher & Powell, 1963) and Langerman (Bersini, Dorigo, Langerman, Seront, & Gambardella, 1996) are highly multimodal, as Ackley and Griewangk, but they are non-symmetrical and their local optima are randomly distributed. In this way, the objective function has no implicit symmetry advantages that might simplify optimization for certain algorithms. Fletcher-Powel function achieves the random distribution of the optima choosing the values of the matrixes **a** and **b**, and of the vector $\alpha$ at random. We have used the values provided by Bäck (1996). For Langerman function, we have used the values of **a** and **c** referenced by Eiben and Bäck (1997b).

## 4. Evolutionary Algorithm

The most suitable evolutionary algorithms to solve optimization problems in continuous domains are evolutionary strategies (Schwefel, 1981; Rechenberg, 1973), genetic algorithms (Holland, 1975; Goldberg, 1989a) with real coding (Goldberg, 1991) and evolutionary programming (Fogel, Owens, & Walsh, 1966; Fogel, 1995). For evaluating CIXL2 we have chosen real coded genetic algorithms, because they are search algorithms of general purpose where the crossover operator plays a central role. The general structure of the genetic algorithm is shown in Figure 4.

Nevertheless, CIXL2 could be applied to any evolutionary algorithms with a crossover or similar operator. On the other hand, the real codification is the most natural one in continuous domains, each gene representing a variable of the function. In this way, the precision of the solution only depends on the data type used to store the variables.

Our objective is the comparison of the behavior of the proposed crossover against other crossovers. This comparison must be made in a common evolutionary framework that is defined by the features of the genetic algorithm. For the definition of such features, we have taken into account the previous studies on the matter. In the following paragraphs we will describe in depth the different components of our genetic algorithm.

### 4.1 Structure of the Individual and Population Size

Each individual is made up of $p = 30$ genes, the dimensionality of the functions to optimize.

The size of the population is one of the critical parameters for many applications. If the size of the population is too small, the algorithm could converge quickly towards suboptimal solutions; if it is too large, too much time and resources could be wasted. It is also

**Genetic algorithm**
**begin**
    $t \leftarrow 0$
    `initialize` $\boldsymbol{\beta}(t)$
    `evaluate` $\boldsymbol{\beta}(t)$
    **while** (**not** `stop-criterion`) **do**
    **begin**
        $t \leftarrow t + 1$
        `select` $\boldsymbol{\beta}(t)$ `from` $\boldsymbol{\beta}(t-1)$
        `crossover` $\boldsymbol{\beta}(t)$
        `mutate` $\boldsymbol{\beta}(t)$
        `evaluate` $\boldsymbol{\beta}(t)$
    **end**
**end**

Figure 4: Structure of the genetic algorithm, $t$ is the current generation.

obvious that the size of the population, together with the selective pressure, influences the diversity of the population.

Several researches have studied these problems from different points of view. Grefenstette (1986) used a meta-genetic algorithm for controlling the parameters of another genetic algorithm, such as population size and the selection method. Goldberg (1989b) made a theoretical analysis of the optimum population size. A study of the influence of the parameters on the search process was carried out by Schaffer, Caruana, Eshelman and Das (1989). Smith (1993) proposed an algorithm that adjusts the size of the population with respect to the error probability of the selection . Another method consists of changing the size of the population (Arabas, Michalewicz, & Mulawka, 1994) dynamically.

The size of the population is usually chosen in an interval between 50 and 500 individuals, depending on the difficulty of the problem. As a general practice, in function optimization, the size is in the interval $[50, 100]$ for unimodal functions, and in the interval $[100, 500]$ for multimodal functions. However, several papers use a compromise size of 100 for all the functions in order to homogenize the comparison environment. We will also use a population size of 100 individuals like other comparative studies (Zhang & Kim, 2000; Takahashi, Kita, & Kobayashi, 1999).

### 4.2 Selection

Zhang and Kim (2000) a comparative study was carried out of the performance of four selection methods: proportional, ranking, tournament and Genitor. In contrast to other studies that are based on an asymptotic study under more or less ideal conditions, this paper is devoted to a practical case, the problem of machine layout. The paper analyzes the quality of the solutions obtained in a reasonable amount of time and using mutation and crossover operators. The study concludes that the methods of ranking and tournament selection obtain better results than the methods of proportional and Genitor selection.

We have chosen the binary tournament selection, against the ranking selection, used by Zhang and Kim (2000) for two reasons:

- The complexity of the tournament selection is lower than the complexity of the ranking selection (Bäck, 1996).
- The selective pressure is higher. This feature allows us to measure whether each crossover is able to keep the population diversity (Goldberg & Deb, 1991).

Tournament selection runs a tournament between two individuals and selects the winner. In order to assure that the best individuals always survive to the next generation, we use elitism, the best individual of the population in generation $t$ is always included in the population in generation $t+1$. It has been proved, both theoretically (Rudolph, 1994) and empirically (Bäck, 1996; Michalewicz, 1992; Zhang & Kim, 2000), the convenience of the use of elitism.

### 4.3 Population Update Model

There are different techniques for updating the population, among the most important are the generational model and the steady-state model. In the generational model in each generation a complete set of $N$ new offspring individuals is created from $N$ parents selected from the population. In most such generational models, the tournament selection is used to choose two parent individuals, and a crossover with $p_c$ probability and a mutation operator con $p_m$ probability are applied to the parents.

This contrasts with the steady-state model, where one member of the population is replaced at a time. The steady-state model selects an individual to be mutated and the mutated individual replaces another individual of the population. For the crossover two individuals are selected and one of the offspring replaces one individual of the population. There are a number of different replacement strategies: replace-worst, replace a randomly chosen member, select replacement using negative fitness.

The model that extrapolates between generational and steady-state is said to have a generation gap $G$ (De Jong, 1975; Jong & Sarma, 1993). Thus for a generational model, $G = 1$; while for a steady-state model, $G = 1/N$. One of the most widely used variants of the steady-stated genetic algorithm is the Minimal Generation Gap (MGG) model (Satoh, Yamamura, & Kobayashi, 1996). This model takes two parents randomly from the population and generates $\lambda$ children. Two individuals are selected from the parents and the offspring: the best individual, and another individual chosen by roulette selection. These two individuals substitute the parents in the population.

The generational model is the most frequently used in the comparative studies that use BLX, SBX, logical crossover and fuzzy recombination. This is the reason why it will be the model used in this paper. However, for UNDX crossover we have used the MGG model, because UNDX and MGG are commonly used together and the generational model can have a negative influence on the performance of UNDX.

For the parameters of the two models we have used the most commonly used in the literature. For the generational model, we use a probability of crossover of $p_c = 0.6$ (De Jong, 1975; Herrera, Lozano, & Verdegay, 1998). For the MGG model we have used $\lambda = 200$, as this is a value commonly used in papers about UNDX (Ono & Kobayashi, 1997; Ono, Kita, & Kobayashi, 1999; Ono, Kobayashi, & Yoshida, 2000). For the mutation probability, values in the interval $p_m \in [0.001, 0.1]$ are usual (De Jong, 1975; Herrera et al., 1998; Michalewicz, 1992; Bäck, 1996). We have chosen a value of $p_m = 0.05$ for both models.

### 4.4 Initialization

In a search algorithm, the initialization method is very important. In many cases the initialization determines the success or failure of the search process. We have opted, as in other papers (Herrera et al., 1998; De Jong, 1975; Beyer & Deb, 2001; Herrera, Lozano, & Sánchez, 2003), for initializing the values of the genes by means of a uniform random distribution within the domain of each variable.

### 4.5 Mutation

As mutation operator we have chosen the non-uniform mutation with parameter $b = 5$ (Michalewicz, 1992) as its dynamical nature makes it very suitable for a wide variety of problems (Herrera & Lozano, 2000).

The individuals $\beta^m$ generated by this mutation are obtained as follows:

$$\beta_i^m = \begin{cases} \beta_i + \triangle(t, b_i - \beta_i) & si \quad \tau = 0 \\ \beta_i - \triangle(t, \beta_i - a_i) & si \quad \tau = 1 \end{cases} \tag{12}$$

being

$$\triangle(t, y) = y(1 - r^{(1-\frac{t}{g_{max}})^b}) \tag{13}$$

where $t$ is the generation, $g_{max}$ is the maximum number of generations, $\tau$ is a random value, $\tau \in \{0, 1\}$, $r$ is a random number in the interval $[0, 1]$ and $b$ is a parameter that determines the degree of dependence of the mutation with regards to the number of iterations. Equation 13 gives values in the interval $[0, y]$. The probability of obtaining a value near 0 increases as the algorithm progresses. This operator performs a uniform search in the initial stages of the evolution, and a very localized search in the final stages.

### 4.6 Stop Criterion

The part of the genetic algorithm that takes up most of the time is the evaluation of the fitness function. The number of evaluations of the fitness in each generation depends on the operators used and the population update model. Different operators and update models can lead to very different numbers of evaluations per generation. That is the reason why it is common to use the number of evaluations as the stop criterion instead of the number of generations. We have used a limit of 300,000 evaluations (Eiben, van der Hauw, & van Hemert, 1998; De Jong & Kosters, 1998) as stop criterion. The precision of the solutions is bounded by the precision of the data type used in the implementation of the genetic algorithm. We have used a double precision data type of 64 bits following the specification ANSI/IEEE STD 754-1985 (IEEE Standard for Binary Floating-Point Arithmetic). This data type has a precision of 15 - 17 digits.

## 5. Analysis of CIXL2

In this section we will perform an analysis of the crossover, and will obtain for every test function the following information:

1. The optimal value for the confidence coefficient $1-\alpha$ of the confidence interval. The values used are $1-\alpha = \{0.70, 0.90, 0.95, 0.99\}$.

2. The optimal number of best individuals used by the crossover to calculate the confidence intervals of the mean. The values used are $n = \{5, 10, 30, 60, 90\}$.

These two factors are not independent, so we will perform an analysis using all the possible pairs $(1-\alpha,\ n)$ of the Cartesian product of the two sets. For each pair we will perform 30 runs of the genetic algorithm with different random seeds. Table 2 shows the average value and standard deviation of the 30 runs for each experiment.

The study of the results has been made by means of an analysis of variance ANOVA II (Dunn & Clark, 1974; Miller, 1981; Snedecor & Cochran, 1980), with the fitness of the best individuals, $A$, as test variable. This fitness is obtained independently in 30 runs and depending on two fixed factors and their interaction. The fixed factors are: the confidence coefficient $C$ with four levels and the number of best individuals $B$ with five levels. The linear model has the form:

$$A_{ij} = \mu + C_i + B_j + CB_{ij} + e_{ij} \qquad (14)$$
$$i = 1,2,3,4; \text{ and } j = 1,2,3,4,5$$

where:

- $C_i$ is the effect over $A$ of the $i$-th level of factor $C$, where $C_1$ represents a confidence coefficient of 0.70, $C_2$ of 0.90, $C_3$ of 0.95 and $C_4$ of 0.99.

- $B_j$ is the effect over $A$ of the $j$-th level of factor $B$, where $B_1$ represents a value of $n = 5$, $B_2$ of $n = 10$, $B_3$ of $n = 30$, $B_4$ of $n = 60$ and $B_5$ of $n = 90$.

- $CB_{ij}$ represents the effect of the interaction between the confidence coefficient $C$ and the number of best individuals $B$.

- $\mu$ is the global mean of the model. The variation of the experimental results from $\mu$ is explained by the effects of the different levels of the factors of the model and their interaction.

- $e_{ij}$ are error variables.

The hypothesis tests try to determine the effect of each term over the fitness of the best individuals, $A$. We have carried out tests for every factor and for the interaction among the factors. This and subsequent tests are performed with a confidence level of 95%. The coefficient $R^2$ of the linear model tells us the percentage of variance of $A$ that is explained by the model.

| Function | n | $1-\alpha$ | Mean | St. Dev. | $1-\alpha$ | Mean | St. Dev. | $1-\alpha$ | Mean | D. Tip. | $1-\alpha$ | Mean | St. Dev. |
|---|---|---|---|---|---|---|---|---|---|---|---|---|---|
| Sphere | 5 | 0.70 | 6.365e-16 | 2.456e-16 | 0.90 | 4.885e-16 | 1.969e-16 | 0.95 | 3.553e-16 | 1.710e-16 | 0.99 | **1.998e-16** | **6.775e-17** |
| $f_{Sph}$ | 10 | | 5.736e-15 | 2.495e-15 | | 2.554e-15 | 8.934e-16 | | 2.642e-15 | 1.258e-15 | | 1.480e-15 | 1.032e-15 |
| | 30 | | 3.728e-12 | 1.623e-12 | | 1.446e-11 | 7.062e-12 | | 2.279e-11 | 1.256e-11 | | 1.248e-10 | 5.914e-11 |
| | 60 | | 6.082e-10 | 2.499e-10 | | 2.867e-08 | 1.642e-08 | | 1.557e-07 | 9.911e-08 | | 5.494e-07 | 6.029e-07 |
| | 90 | | 3.838e-09 | 2.326e-09 | | 4.383e-08 | 3.068e-08 | | 6.840e-08 | 5.894e-08 | | 1.061e-07 | 8.401e-08 |
| Schwefel's | 5 | 0.70 | **1.995e-03** | **2.280e-03** | 0.90 | 8.403e-03 | 7.748e-03 | 0.95 | 7.662e-03 | 9.693e-03 | 0.99 | 1.305e-02 | 1.303e-02 |
| double sum | 10 | | 2.232e-02 | 2.859e-02 | | 5.407e-02 | 3.792e-02 | | 4.168e-02 | 4.383e-02 | | 1.462e-02 | 1.422e-02 |
| $f_{SchDS}$ | 30 | | 8.464e-02 | 1.168e-01 | | 3.190e-01 | 2.798e-01 | | 2.644e-01 | 2.569e-01 | | 1.223e-01 | 9.018e-02 |
| | 60 | | 1.376e-01 | 1.202e-01 | | 4.059e-01 | 2.395e-01 | | 2.223e-01 | 1.384e-01 | | 2.134e-01 | 1.464e-01 |
| | 90 | | 8.048e-01 | 5.403e-01 | | 2.257e+00 | 1.490e+00 | | 7.048e-01 | 7.689e-01 | | 2.799e-01 | 2.322e-01 |
| Rosenbrock | 5 | 0.70 | 2.494e+01 | 1.283e+00 | 0.90 | 2.506e+01 | 3.050e-01 | 0.95 | 2.497e+01 | 4.663e-01 | 0.99 | **2.463e+01** | **1.330e+00** |
| $f_{Ros}$ | 10 | | 2.579e+01 | 2.044e-01 | | 2.591e+01 | 1.324e-01 | | 2.589e+01 | 9.426e-02 | | 2.579e+01 | 1.609e-01 |
| | 30 | | 2.611e+01 | 1.471e-01 | | 2.632e+01 | 1.745e-01 | | 2.642e+01 | 1.377e-01 | | 2.668e+01 | 9.999e-02 |
| | 60 | | 2.576e+01 | 1.988e-01 | | 2.593e+01 | 2.292e-01 | | 2.600e+01 | 4.045e-01 | | 2.617e+01 | 4.787e-01 |
| | 90 | | 2.562e+01 | 2.827e-01 | | 2.570e+01 | 2.974e-01 | | 2.579e+01 | 2.629e-01 | | 2.585e+01 | 3.654e-01 |
| Rastrigin | 5 | 0.70 | **2.919e+00** | **1.809e+00** | 0.90 | 6.036e+00 | 2.023e+00 | 0.95 | 7.893e+00 | 2.450e+00 | 0.99 | 7.164e+00 | 2.579e+00 |
| $f_{Ras}$ | 10 | | 6.799e+00 | 2.480e+00 | | 1.068e+01 | 3.786e+00 | | 1.297e+01 | 3.844e+00 | | 1.675e+01 | 6.554e+00 |
| | 30 | | 9.452e+00 | 2.434e+00 | | 1.270e+01 | 3.522e+00 | | 1.327e+01 | 4.770e+00 | | 1.552e+01 | 3.664e+00 |
| | 60 | | 1.413e+01 | 4.126e+00 | | 1.837e+01 | 6.070e+00 | | 1.499e+01 | 4.434e+00 | | 1.691e+01 | 4.123e+00 |
| | 90 | | 1.771e+01 | 5.063e+00 | | 2.438e+01 | 7.688e+00 | | 1.987e+01 | 5.637e+00 | | 2.249e+01 | 6.058e+00 |
| Schwefel | 5 | 0.70 | **6.410e+02** | **2.544e+02** | 0.90 | 1.145e+03 | 5.422e+02 | 0.95 | 1.424e+03 | 6.837e+02 | 0.99 | 2.844e+03 | 4.168e+02 |
| $f_{Sch}$ | 10 | | 1.793e+03 | 4.172e+02 | | 1.325e+03 | 2.340e+02 | | 1.486e+03 | 2.607e+02 | | 2.525e+03 | 3.069e+02 |
| | 30 | | 2.675e+03 | 2.592e+02 | | 2.264e+03 | 2.758e+02 | | 2.061e+03 | 2.369e+02 | | 1.986e+03 | 2.424e+02 |
| | 60 | | 2.700e+03 | 1.471e+02 | | 2.513e+03 | 1.927e+02 | | 2.496e+03 | 2.146e+02 | | 2.169e+03 | 2.434e+02 |
| | 90 | | 2.738e+03 | 1.476e+02 | | 2.704e+03 | 1.516e+02 | | 2.672e+03 | 1.349e+02 | | 2.529e+03 | 1.837e+02 |
| Ackley | 5 | 0.70 | 1.378e-08 | 5.677e-09 | 0.90 | 6.320e-09 | 2.966e-09 | 0.95 | **4.677e-09** | **1.960e-09** | 0.99 | 5.188e-09 | 2.883e-09 |
| $f_{Ack}$ | 10 | | 2.074e-07 | 9.033e-08 | | 9.544e-08 | 3.422e-08 | | 9.396e-08 | 3.513e-08 | | 5.806e-08 | 2.683e-08 |
| | 30 | | 8.328e-06 | 1.403e-06 | | 1.483e-05 | 3.956e-06 | | 2.246e-05 | 4.957e-06 | | 4.976e-05 | 1.298e-05 |
| | 60 | | 1.019e-04 | 2.396e-05 | | 8.292e-04 | 2.097e-04 | | 1.897e-03 | 9.190e-04 | | 3.204e-03 | 1.373e-03 |
| | 90 | | 2.518e-04 | 7.167e-05 | | 7.544e-04 | 2.668e-04 | | 9.571e-02 | 3.609e-01 | | 1.741e-01 | 5.290e-01 |
| Griewangk | 5 | 0.70 | 1.525e-02 | 1.387e-02 | 0.90 | 2.463e-02 | 2.570e-02 | 0.95 | 1.574e-02 | 1.411e-02 | 0.99 | 1.285e-02 | 1.801e-02 |
| $f_{Gri}$ | 10 | | 1.647e-02 | 1.951e-02 | | 2.695e-02 | 2.713e-02 | | 2.195e-02 | 2.248e-02 | | 3.194e-02 | 3.680e-02 |
| | 30 | | 2.012e-02 | 2.372e-02 | | 1.819e-02 | 1.664e-02 | | 2.321e-02 | 3.842e-02 | | 2.254e-02 | 1.877e-02 |
| | 60 | | 7.884e-03 | 1.061e-02 | | 2.808e-02 | 9.686e-02 | | 7.410e-03 | 1.321e-02 | | 1.582e-02 | 2.727e-02 |
| | 90 | | 7.391e-03 | 7.617e-03 | | **5.248e-03** | **6.741e-03** | | 8.938e-03 | 1.196e-02 | | 1.230e-02 | 2.356e-02 |
| Fletcher | 5 | 0.70 | 1.523e+04 | 1.506e+04 | 0.90 | 2.293e+04 | 1.882e+04 | 0.95 | **1.286e+04** | **1.317e+04** | 0.99 | 1.527e+04 | 1.362e+04 |
| $f_{Fle}$ | 10 | | 1.966e+04 | 1.585e+04 | | 2.248e+04 | 2.300e+04 | | 1.633e+04 | 1.344e+04 | | 1.891e+04 | 1.612e+04 |
| | 30 | | 2.145e+04 | 1.631e+04 | | 2.129e+04 | 1.310e+04 | | 3.049e+04 | 2.306e+04 | | 2.492e+04 | 1.967e+04 |
| | 60 | | 2.133e+04 | 2.110e+04 | | 2.124e+04 | 1.213e+04 | | 2.935e+04 | 2.155e+04 | | 2.374e+04 | 1.479e+04 |
| | 90 | | 2.432e+04 | 2.273e+04 | | 2.898e+04 | 3.131e+04 | | 2.918e+04 | 2.418e+04 | | 3.453e+04 | 2.498e+04 |
| Langerman | 5 | 0.70 | -2.064e-01 | 9.346e-02 | 0.90 | -2.544e-01 | 1.401e-01 | 0.95 | **-3.545e-01** | **1.802e-01** | 0.99 | -2.803e-01 | 1.350e-01 |
| $f_{Lan}$ | 10 | | -2.339e-01 | 1.280e-01 | | -2.582e-01 | 1.574e-01 | | -2.663e-01 | 1.247e-01 | | -2.830e-01 | 1.645e-01 |
| | 30 | | -2.124e-01 | 1.038e-01 | | -2.191e-01 | 1.100e-01 | | -1.908e-01 | 9.776e-02 | | -2.382e-01 | 1.572e-01 |
| | 60 | | -1.975e-01 | 1.405e-01 | | -1.752e-01 | 7.145e-02 | | -1.762e-01 | 8.929e-02 | | -1.949e-01 | 9.500e-02 |
| | 90 | | -1.599e-01 | 9.057e-02 | | -1.336e-01 | 6.042e-02 | | -1.656e-01 | 8.336e-02 | | -1.796e-01 | 8.453e-02 |

Table 2: Average value and standard deviation of the 30 runs for each experiment

For determining whether there are significant differences among the various levels of a factor, we perform a multiple comparison test of the average fitness obtained with the different levels of each factor. First, we carry out a Levene test (Miller, 1996; Levene, 1960) for evaluating the equality of variances. If the hypothesis that the variances are equal is accepted, we perform a Bonferroni test (Miller, 1996) for ranking the means of each level of the factor. Our aim is to find the level of each factor whose average fitness is significantly better than the average fitness of the rest of the levels of the factor. If the test of Levene results in rejecting the equality of covariance matrixes, we perform a Tamhane test (Tamhane & Dunlop, 2000) instead of a Bonferroni test. Tables 9, 12, and 13 in Appendix A show the results obtained following the above methodology.

For Sphere function, the significant levels $\alpha^*$ of each term of the linear model on Table 9 show that none of the factors of the linear model has a significant effect on the model built to explain the variance of the fitness $A$. This effect is due to the fact that $f_{Sph}$ is easy to optimize and the fitness behaves as a singular random variable with sample variance near 0. We can see in Table 2 that the best results are obtained with the pair $(0.99, 5)$. The multiple comparison test of Table 12 confirms that the means obtained with the value $n = 5$ are significatively better than the means obtained with other values. In the same way, the average fitness for $1 - \alpha = 0.70$ is significantly the best one. The results show that, for any value of $n$, the best value of $1 - \alpha$, in general, is $1 - \alpha = 0.70$. Due to the simple form of $f_{Sph}$, the best parameters of the crossover show a high exploitative component with a fast shifting towards the region of the best individuals.

For the unimodal and non-separable functions $f_{SchDS}$ and $f_{Ros}$, both factors and their interaction are significant in the linear model that explains the sample variance of $A$ with a determination coefficient around 0.5. Table 2 shows that the best results are obtained with $n = 5$; the Tamhane test shows that the means obtained with this value of $n$ are significatively better than the means obtained with other values. The results for the value of the confidence coefficient are less conclusive. In fact, for $f_{Ros}$ there are no significant differences among the different values of $1 - \alpha$, although the best results are obtained with $1 - \alpha = 0.7$. For $f_{SchDS}$ the average fitness for $\mu_{0.99}$ is the best one, but without significant differences with $\mu_{0.70}$. $\mu_{0.70}$ together with $n = 5$ is the one that shows the best results. We can conclude that the feature of non-separability of the functions does not imply a notable change in the parameters of the crossover with respect to the parameters used for $f_{Sph}$.

For $f_{Ras}$ and $f_{Sch}$, which are separable and multimodal, the most adequate pair of parameters is $(0.70, 5)$. For $f_{Ras}$, the test shows that the performance of this pair is significantly better. However, for $f_{Sch}$, the best mean is obtained with $\mu_5$ with results that are significantly better than these obtained with other values, with the exception of $\mu_{10}$. There are no significant differences among $\mu_{0.70}$, $\mu_{0.95}$ and $\mu_{90}$. The three factors of the linear model are significant with quite large determination coefficients of 0.617 for $f_{Ras}$ and 0.805 for$f_{Sch}$. This means that the factors and their interaction explain a high percentage of the variance of the fitness $A$.

For $f_{Ack}$, the best results are obtained with the pair $(0.95, 5)$. The Tamhane test confirms that $n = 5$ is the most suitable value, while there are no significant differences among $\mu_{0.70}$, $\mu_{0.95}$ and $\mu_{0.99}$. For $f_{Gri}$ the best results are obtained with the pair $(0.90, 90)$. The test shows that large values of $n$ are the most suitable for the optimization of this function. There are no significant differences among the performance of the different values of $1 - \alpha$.

For both functions the determination coefficient of the linear model is low, showing that the linear model does not explain the variance of the fitness. The lack of a linear relation among $n$, $1-\alpha$ and the fitness makes it more difficult to determine the best value of the parameters of the crossover.

The case of $f_{Fle}$ and $f_{Lan}$ is similar, as the linear model hardly gives any information about the effect of the parameters on the fitness. The most adequate pair for the optimization of these two functions is $(0.95, 5)$. The test shows that the best values of $n$ are $n = 5$ and $n = 10$. On the other hand, there are no significant differences among the performance of the crossover for the different values of $1 - \alpha$.

The overall results show that the selection of the best $n = 5$ individuals of the population would suffice for obtaining a localization estimator good enough to guide the search process even for multimodal functions where a small value of $n$ could favor the convergence to local optima. However, if the virtual parents have a worse fitness than the parent from the population, the offspring is generated near the latter, and the domain can be explored in multiple directions. In this way, the premature convergence to suboptimal virtual parents is avoided.

However, if the best $n$ individuals are concentrated in a local optimum the algorithm will very likely converge to such optimum. That is the reason why in complex functions a larger value of $n$ may be reasonable, adding to the confidence interval individuals located in or near different optima. As an example of this, the case of $f_{Gri}$ for which the best results are achieved with $n = 90$ and $n = 60$ is noteworthy.

The confidence coefficient bounds the error in the determination of the localization parameter and is responsible for focussing the search. The multiple comparison tests show that the value $1 - \alpha = 0.70$ is the best for 6 problems, and is, as least, no worse than the best one in the other problems. So it can be chosen as the most adequate value of the parameter.

## 6. Comparative Study of the Crossovers

Due to the large amount of different crossovers available, it is unfeasible to make a comprehensive comparison between all those crossovers and CIXL2. We have chosen those crossovers that obtain interesting results and those whose features are similar to our crossover, that is, which are self-adaptive and establish a balance between the exploration and the exploitation of the search space. The way in which these two features are balanced is regulated by one or more parameters of each crossover. These parameters have been chosen following the authors' recommendations and the papers devoted to the comparison of the different operators.

The crossovers used in the comparison are: BLX$\alpha$ (Eshelman & Schaffer, 1993) with different degrees of exploration determined by the values $\alpha = \{0.2, 0.5\}$ (Herrera et al., 2003); fuzzy recombination (Voigt et al., 1995); based on fuzzy connectives of the logical family (logical crossover) (Herrera et al., 1998) using S2 strategies and $\lambda = 0.5$ (Herrera & Lozano, 2000), SBX (Deb & Agrawal, 1995) using the values $\nu = \{2, 5\}$ (Deb & Beyer, 2001); UNDX (Ono & Kobayashi, 1997) with $\sigma_\xi = \frac{1}{2}$ and $\sigma_\eta = \frac{0.35}{\sqrt{p}}$ (Kita, Ono, & Kobayashi, 1998; Kita, 2001). For CIXL2, as we have determined in the previous study, we will use $n = 5$ and $1 - \alpha = 0.70$.

Following the setup of the previous study, we performed an ANOVA II analysis and a multiple comparison test. As might have been expected, keeping in mind the "no-free lunch" theorem and the diversity of the functions of the test set, the tests show that there is no crossover whose results are significatively better than the results of all other crossovers. This does not mean that these differences could not exist for certain kinds of functions. So, in order to determine for each kind of function whether a crossover is better than the others, we have performed an ANOVA I analysis — where the only factor is the crossover operator — and a multiple comparison test. Additionally, we graphically study the speed of convergence of the RCGA with regard to the crossover operator. In order to enforce the clearness of the graphics for each crossover, we show only the curve of the best performing set of parameters for BLX and SBX crossovers.

| Crossover | Mean | St.Dev. | Mean | St.Dev. | Mean | St.Dev. |
|---|---|---|---|---|---|---|
| | $\mathbf{f_{Sph}}$ | | $\mathbf{f_{SchDS}}$ | | $\mathbf{f_{Ros}}$ | |
| CIXL2 | 6.365e-16 | 2.456e-16 | **1.995e-03** | 2.280e-03 | **2.494e+01** | 1.283e+00 |
| BLX(0.3) | **3.257e-16** | 1.396e-16 | 1.783e-02 | 1.514e-02 | 2.923e+01 | 1.723e+01 |
| BLX(0.5) | 4.737e-16 | 4.737e-16 | 9.332e-03 | 1.086e-02 | 3.161e+01 | 2.094e+01 |
| SBX(2) | 1.645e-12 | 8.874e-13 | 2.033e-01 | 1.966e-01 | 2.775e+01 | 9.178e+00 |
| SBX(5) | 4.873e-12 | 3.053e-12 | 3.933e-01 | 2.881e-01 | 3.111e+01 | 1.971e+01 |
| Ext. F. | 2.739e-15 | 1.880e-15 | 3.968e+01 | 1.760e+01 | 2.743e+01 | 1.394e+01 |
| Logical | 3.695e-13 | 1.670e-13 | 1.099e+01 | 7.335e+00 | 2.703e+01 | 8.358e-02 |
| UNDX | 2.910e-05 | 1.473e-05 | 2.080e+01 | 7.216e+00 | 2.840e+01 | 3.606e-01 |
| | $\mathbf{f_{Ras}}$ | | $\mathbf{f_{Sch}}$ | | $\mathbf{f_{Ack}}$ | |
| CIXL2 | 2.919e+00 | 1.809e+00 | 6.410e+02 | 2.544e+02 | **1.378e-08** | 5.677e-09 |
| BLX(0.3) | **2.189e+00** | 1.417e+00 | **3.695e+02** | 1.595e+02 | 4.207e-08 | 1.713e-08 |
| BLX(0.5) | 3.018e+00 | 1.683e+00 | 4.200e+02 | 1.916e+02 | 6.468e-08 | 1.928e-08 |
| SBX(2) | 1.844e+01 | 4.417e+00 | 1.470e+03 | 3.827e+02 | 5.335e-06 | 1.453e-06 |
| SBX(5) | 1.419e+01 | 3.704e+00 | 1.104e+03 | 3.353e+02 | 9.662e-06 | 2.377e-06 |
| Ext. F. | 2.245e+01 | 4.914e+00 | 3.049e+03 | 2.876e+02 | 1.797e-07 | 5.823e-08 |
| Logical | 6.325e+01 | 1.012e+01 | 2.629e+03 | 9.749e+01 | 2.531e-06 | 7.129e-07 |
| UNDX | 1.107e+02 | 1.242e+01 | 8.050e+03 | 3.741e+02 | 3.551e-02 | 1.224e-02 |
| | $\mathbf{f_{Gri}}$ | | $\mathbf{f_{Fle}}$ | | $\mathbf{f_{Lan}}$ | |
| CIXL2 | 1.525e-02 | 1.387e-02 | **1.523e+04** | 1.506e+04 | -2.064e-01 | 9.346e-02 |
| BLX(0.3) | 4.749e-02 | 4.579e-02 | 1.570e+04 | 1.515e+04 | -3.003e-01 | 1.388e-01 |
| BLX(0.5) | 3.760e-02 | 2.874e-02 | 1.802e+04 | 1.483e+04 | **-3.457e-01** | 1.684e-01 |
| SBX(2) | 2.196e-02 | 1.874e-02 | 3.263e+04 | 3.110e+04 | -1.939e-01 | 1.086e-01 |
| SBX(5) | 3.128e-02 | 2.737e-02 | 3.333e+04 | 2.973e+04 | -1.866e-01 | 9.080e-02 |
| Ext. F. | **1.315e-03** | 3.470e-03 | 1.691e+04 | 1.446e+04 | -1.064e-01 | 5.517e-02 |
| Logical | 6.078e-03 | 6.457e-03 | 2.718e+04 | 1.388e+04 | -7.396e-08 | 2.218e-07 |
| UNDX | 7.837e-02 | 4.438e-02 | 3.469e+04 | 2.136e+04 | -2.130e-01 | 9.116e-02 |

Table 3: Average values and standard deviation for the 30 runs of every crossover operator.

Table 3 shows the average values and standard deviations for the 30 runs performed for each crossover operator. Table 10 in Appendix A shows how, for all the functions, except $f_{Ros}$, the crossover operator has a significant effect on the linear model. The table also shows that the results of the Levene test indicate the inequality of the variances of the results of all the functions, excepting $f_{Fle}$. So, we use the Bonferroni test for $f_{Fle}$, and the Tamhane test for all the others. The results of the multiple comparison test, the ranking established by the tests and the significant level of the differences among the results of the crossovers are shown on Tables 14, 15 and 16 (Appendix A). Figures 5 - 13, in Appendix B, show, in logarithmic scale, the convergence curves for each function.

For $f_{Sph}$ the high value of the determination coefficient shows that the linear model explains much of the variance of the fitness. The best values are obtained with BLX(0.3), BLX(0.5) and CIXL2, in this order. With these operators we obtain precisions around 1e-16. Figure 5 shows that CIXL2 is the fastest in convergence, but it is surpassed by BLX in the last generations.

For $f_{SchDS}$ and $f_{Ros}$ the best results are obtained with CIXL2. For $f_{SchDS}$ the difference in performance with the other crossovers is statistically significant. For $f_{Ros}$ the differences are significant, when CIXL2 is compared with Logical and UNDX. For $f_{SchDS}$ the Figure 6 shows how CIXL2 achieves a quasi-exponential convergence and a more precise final result. For $f_{Ros}$, in the Figure 7 we can see how the speed of convergence of CIXL2 is the highest, although the profile of all the crossovers is very similar with a fast initial convergence followed by a poor evolution due to the high epistasis of the function. The differences in the overall process are small. This fact explains that in the linear model the influence of the factor crossover is not significant and the determination coefficient is small.

For $f_{Ras}$, BLX(0.3) again obtains the best results but without significant difference to the average values obtained with CIXL2 and BLX(0.5). These three operators also obtain the best results for $f_{Sch}$; however, the tests show that there are significant differences between CIXL2 and BLX(0.5), and that there are no differences between BLX(0.5) and BLX(0.3). The latter obtains the best results. Figures 8 and 9 show that BLX is the best in terms of convergence speed followed by CIXL2. The large value of $R^2$ means that the crossover has a significant influence on the evolutive process.

For $f_{Ack}$, CIXL2 obtains significantly better results. In Figure 10 we can see how it also converges faster. The large value of $R^2$ means that the crossover has a significant influence on the evolutive process. For $f_{Gri}$, the Fuzzy operator obtains significantly better results. The following ones, with significant differences between them, are Logical and CIXL2. Figure 11 shows a fast initial convergence of CIXL2, but in the end Logical and Fuzzy obtain better results.

For $f_{Fle}$ the best results are obtained with CIXL2, but the difference is only significant with SBX and UNDX. Figure 12 shows that CIXL2 is the fastest in convergence, but with a curve profile similar to BLX and Fuzzy. For $f_{Lan}$, the best operator is BLX(0.5), with differences that are significant for all the other operators with the exception of BLX(0.3). UNDX and CIXL2 are together in third place. Figure 13 shows that the behavior of all crossovers is similar, except for the Logical crossover that converges to a value far from the other operators.

## 7. Comparison with Estimation of Distribution Algorithms

EDAs are evolutionary algorithms that use, as CIXL2, the best individuals of the population to direct the search. A comparison with this paradigm is interesting, although there are significant differences between EDAs and RCGAs.

EDAs remove the operators of crossover and mutation. In each generation a subset of the population is selected and the distribution of the individuals of this subset is estimated. The individuals of the population for the next generation are obtained sampling the estimated distribution. Although any selection method could be applied, the most common one is the selection of the best individuals of the population.

The first EDAs were developed for discrete spaces. Later, they were adapted to continuous domains. We can distinguish two types of EDAs, whether they take into account dependencies between the variables or not. One of the most used among the EDAs that do not consider dependencies is $UMDA_c$ (Univariate Marginal Distribution Algorithm for continuous domains) (Larrañaga, Etxeberria, Lozano, & Peña, 2000). In every generation and for every variable the $UMDA_c$ carries out some statistical test in order to find the density function that best fits the variable. Once the densities have been identified, the estimation of parameters is performed by their maximum likelihood estimates. If all the distributions are normal, the two parameters are the mean and the standard deviation. This particular case will be denoted $UMDA_c^G$ (Univariate Marginal Distribution Algorithm for Gaussian models).

Among the other type of EDAs, we can consider $EGNA_{BGe}$ (Estimation of Gaussian Network Algorithm) (Larrañaga et al., 2000) whose good results in function optimization are reported by Bengoetxea and Miquélez (2002). In each generation, $EGNA_{BGe}$ learns the Gaussian network structure by using a Bayesian score that gives the same value for Gaussian networks reflecting the same conditional dependencies are used. Next, it calculates estimations for the parameters of the Gaussian network structure.

In the experiments we have used the parameters reported by Bengoetxea and T. Miquélez (2002): a population of 2000 individuals, initialized using a uniform distribution, from which a subset of the best 1000 individuals are selected to estimate the density function, and the elitist approach was chosen (the best individual is included for the next population and 1999 individuals are simulated). Each algorithm has been run 30 times with a stop criterion of 300,000 evaluations of the fitness function.

The results of EDAs are compared with the results of a RCGA with CIXL2 of parameters $n = 5$ and $1 - \alpha = 0.70$. We performed an ANOVA I analysis where the three levels of the factor are the different algorithms: RCGA with CIXL2, $UMDA_c$ and $EGNA_{BGe}$. We also carried out a multiple comparison test.

Table 4 shows the average values and standard deviations for 30 runs for each algorithm. Table 11 in Appendix A shows how, for all the functions excepting $f_{Ack}$, the type of algorithm has a significant effect over the linear model and exist inequality of the variances of the results (Levene test). So, we have used Tamhane test for all the functions and Bonferroni test for $f_{Ack}$. Table 17 (Appendix A) shows the results of the multiple comparison test and the ranking established by the test.

For $f_{Sph}$ the results are very similar. The fitness behaves as a singular random variable with sample variance near 0 and the statistical tests are not feasible.

For $f_{SchDS}$ the results of CIXL2 are significantly better than the results of $UMDA_c$ and $EGNA_{BGe}$. The same situation occurs for $f_{Ros}$, $f_{Ras}$, $f_{Sch}$ and $f_{Ack}$, with the exception that in these four functions there are no significant differences between the two EDAs. For $f_{Gri}$, $EGNA_{BGe}$ and $UMDA_c$ achieve the best results, significantly better than CIXL2. For $f_{Fle}$, $UMDA_c$ is significantly better than $EGNA_{BGe}$ and CIXL2, but there are no differences between these two. For $f_{Lan}$, CIXL2 obtains the best results, but there are no significant differences among the three algorithms.

The estimation of the distribution function of the best individuals of the population performed by EDAs is an advantage in $f_{Sph}$, unimodal and separable, and $f_{Gri}$ and $f_{Ack}$ whose optima are regularly distributed. The results of EDAs for $f_{Gri}$ are better than the results of CIXL2, but the results for $f_{Ack}$ are worse. The results for $f_{Sph}$ of all the algorithms are very similar. For non-separable unimodal functions, such as $f_{SchDS}$ and $f_{Ros}$, the interdependence among their variables should favor the performance of $EGNA_{BGe}$ over $UMDA_c$ and CIXL2. Nevertheless, CIXL2 achieves the best results for these two functions. For multimodal separable functions, $f_{Ras}$ and $f_{Sch}$, it is difficult to identify the distribution of the best individuals and the performance of EDAs is below the performance of CIXL2.

For extremely complex functions, such as $f_{Fle}$ and $f_{Lan}$, the results are less conclusive. For $f_{Fle}$ the best results are obtained with $UMDA_c$, and there are no differences between $EGNA_{BGe}$ and CIXL2. For $f_{Lan}$, CIXL2 achieves the best results, but the differences among the three algorithms are not statistically significant.

| EA | Mean | St.Dev. | Mean | St.Dev. | Mean | St.Dev. |
|---|---|---|---|---|---|---|
| | $\mathbf{f_{Sph}}$ | | $\mathbf{f_{SchDS}}$ | | $\mathbf{f_{Ros}}$ | |
| $CIXL2$ | 6.365e-16 | 2.456e-16 | **1.995e-03** | 2.280e-03 | **2.494e+01** | 1.283e+00 |
| $UMDA_c$ | 1.196e-16 | 1.713e-17 | 2.221e+01 | 3.900e+00 | 2.787e+01 | 2.278e-02 |
| $EGNA_{BGe}$ | **1.077e-16** | 1.001e-17 | 2.096e-01 | 1.189e-01 | 2.785e+01 | 1.629e-01 |
| | $\mathbf{f_{Ras}}$ | | $\mathbf{f_{Sch}}$ | | $\mathbf{f_{Ack}}$ | |
| CIXL2 | **2.919e+00** | 1.809e+00 | **6.410e+02** | 2.544e+02 | **1.378e-08** | 5.677e-09 |
| $UMDA_c$ | 1.576e+02 | 7.382e+00 | 1.153e+04 | 9.167e+01 | 2.478e-08 | 1.831e-09 |
| $EGNA_{BGe}$ | 1.563e+02 | 8.525e+00 | 1.155e+04 | 8.754e+01 | 2.297e-08 | 2.095e-09 |
| | $\mathbf{f_{Gri}}$ | | $\mathbf{f_{Fle}}$ | | $\mathbf{f_{Lan}}$ | |
| CIXL2 | 1.525e-02 | 1.387e-02 | 1.523e+04 | 1.506e+04 | **-2.064e-01** | 9.346e-02 |
| $UMDA_c$ | 9.465e-16 | 1.207e-16 | **5.423e+03** | 1.562e+03 | -1.734e-01 | 4.258e-11 |
| $EGNA_{BGe}$ | **8.200e-16** | 1.149e-16 | 9.069e+03 | 7.592e+03 | -1.734e-01 | 1.864e-11 |

Table 4: Average values and standard deviation for the 30 runs of three evolutionary algorithms: RCGA with CIXL2 crossover, $UMDA_c$ and $EGNA_{BGe}$.

## 8. Application to Artificial Intelligence

Genetic algorithms have been applied to almost any kind of problem, such as, object recognition for artificial vision (Singh, Chatterjee, & Chaudhury, 1997; Bebis, Louis, Varol, & Yfantis, 2002), robotics path planing (Davidor, 1991; Sedighi, Ashenayi, Manikas, Wainwright, & Tai, 2004), parameter estimation (Johnson & Husbands, 1990; Ortiz-Boyer, Hervás-Martínez, & Muñoz-Pérez, 2003), instance selection (Cano, Herrera, & Lozano, 2003; Kuncheva, 1995), reinforcement learning (Moriarty, Schultz, & Grefenstette, 1999), and neural network (Miller, Todd, & Hedge, 1991; Andersen & Tsoi, 1993; Bebis, Georgiopoulos, & Kasparis, 1997) and ensemble design (Zhou, Wu, & Tang, 2002).

Real-coded genetic algorithms using CIXL2 can be applied to any of these problems provided they are defined in a continuous domain. We have chosen an application of RCGAs to the estimation of the weight of each network in an ensemble. This is an interesting problem where standard methods encounter many difficulties.

### 8.1 Estimation of the Weights of the Networks of an Ensemble

Neural network ensembles (Perrone & Cooper, 1993) (García-Pedrajas, Hervás-Martínez, & Ortiz-Boyer, 2005) are receiving increasing attention in recent neural network research, due to their interesting features. They are a powerful tool specially when facing complex problems. Network ensembles are made up of a linear combination of several networks that have been trained using the same data, although the actual sample used by each network to learn can be different. Each network within the ensemble has a potentially different weight in the output of the ensemble. Several papers have shown (Perrone & Cooper, 1993) that the network ensemble has a generalization error generally smaller than that obtained with a single network and also that the variance of the ensemble is lesser than the variance of a single network. The output of an ensemble, $y$, when an input pattern $\mathbf{x}$ is presented, is:

$$y(\mathbf{x}) = \sum_{i=1}^{k} \alpha_i y_i(\mathbf{x}), \tag{15}$$

where $y_i$ is the output of network $i$, and $w_i$ is the weight associated to that network. If the networks have more than one output, a different weight is usually assigned to each output. The ensembles of neural networks have some of the advantages of large networks without their problems of long training time and risk of over-fitting.

Moreover, this combination of several networks that cooperate in solving a given task has other important advantages, such as (Liu, Yao, & Higuchi, 2000; Sharkey, 1996):

- They can perform more complex tasks than any of their subcomponents.
- They can make an overall system easier to understand and modify.
- They are more robust than a single network.

Techniques using multiple models usually consist of two independent phases: model generation and model combination (Merz, 1999b). Once each network has been trained and assigned a weights (model generation), there are, in a classification environment three basic methods for combining the outputs of the networks (model combination):

1. *Majority voting.* Each pattern is classified into the class where the majority of networks places it (Merz, 1999b). Majority voting is effective, but is prone to fail in two scenarios:
   (a) When a subset of redundant and less accurate models comprise the majority, and
   (b) When a dissenting vote is not recognized as an area of specialization for a particular model.
2. *Sum of the outputs of the networks.* The output of the ensemble is just the sum of the outputs of the individual networks.
3. *Winner takes all.* The pattern is assigned to the class with the highest output over all the outputs of all the networks. That is, the network with the largest outputs directly classify the pattern, without taking into account the other networks.

The most commonly used methods for combining the networks are the *majority voting* and *sum of the outputs of the networks*, both with a weight vector that measures the confidence in the prediction of each network. The problem of obtaining the weight vector $\boldsymbol{\alpha}$ is not an easy task. Usually, the values of the weights $\alpha_i$ are constrained:

$$\sum_{i=1}^{N} \alpha_i = 1, \tag{16}$$

in order to help to produce estimators with lower prediction error (Leblanc & Tibshirani, 1993), although the justification of this constraint is just intuitive (Breiman, 1996). When the method of majority voting is applied, the vote of each network is weighted before it is counted:

$$F(\boldsymbol{x}) = \arg\max_y \sum_{i:f_i(\boldsymbol{x})=y} \alpha_i. \tag{17}$$

The problem of finding the *optimal* weight vector is a very complex task. The "Basic ensemble method (BEM)", as it is called by Perrone and Cooper (1993), consists of weighting all the networks equally. So, having $N$ networks, the output of the ensembles is:

$$F(\boldsymbol{x}) = \frac{1}{N} \sum_{i=1}^{N} f_i(\boldsymbol{x}). \tag{18}$$

Perrone and Cooper (1993) defined the *Generalized Ensemble Method*, which is equivalent to the *Mean Square Error - Optimal Linear Combination* (MSE-OLC) without a constant term of Hashem (Hashem, 1997). The form of the output of the ensemble is:

$$f_{GEM}(\boldsymbol{x}) \equiv \sum_{i=1}^{N} \alpha_i f_i(\boldsymbol{x}), \tag{19}$$

where the $\alpha_i' s$ are real and satisfy the constraint $\sum_{i=1}^{N} \alpha_i = 1$. The values of $\alpha_i$ are given by:

$$\alpha_i = \frac{\sum_j C_{ij}^{-1}}{\sum_k \sum_j C_{kj}^{-1}}. \tag{20}$$

where $C_{ij}$ is the symmetric correlation matrix $C_{ij} \equiv E[m_i(\boldsymbol{x}) m_j(\boldsymbol{x})]$, where $m_k(\boldsymbol{x})$ defines the *misfit* of function $k$, that is the deviation from the true solution $f(\boldsymbol{x})$, $m_k(\boldsymbol{x}) \equiv f(\boldsymbol{x}) - f_k(\boldsymbol{x})$. The previous methods are commonly used. Nevertheless, many other techniques have been proposed over the last few years. Among others, there are methods based on linear regression (Leblanc & Tibshirani, 1993), principal components analysis and least-square regression (Merz, 1999a), correspondence analysis (Merz, 1999b), and the use of a validation set (Opitz & Shavlik, 1996).

In this application, we use a genetic algorithm for obtaining the weight of each component. This approach is similar to the use of a gradient descent procedure (Kivinen & Warmuth, 1997), avoiding the problem of being trapped in local minima. The use of a genetic algorithm has an additional advantage over the optimal linear combination, as the former is not affected by the collinearity problem (Perrone & Cooper, 1993; Hashem, 1997).

### 8.1.1 Experimental Setup

Each set of available data was divided into two subsets: 75% of the patterns were used for learning, and the remaining 25% for testing the generalization of the networks. There are two exceptions, Sonar and Vowel problems, as the patterns of these two problems are prearranged in two specific subsets due to their particular features. A summary of these data sets is shown in Table 5. No validation set was used in our experiments.

| Data set | Cases | | Classes | Features | | | Inputs |
|---|---|---|---|---|---|---|---|
| | Train | Test | | C | B | N | |
| Anneal | 674 | 224 | 5 | 6 | 14 | 18 | 59 |
| Autos | 154 | 51 | 6 | 15 | 4 | 6 | 72 |
| Balance | 469 | 156 | 3 | 4 | – | – | 4 |
| Breast-cancer | 215 | 71 | 2 | – | 3 | 6 | 15 |
| Card | 518 | 172 | 2 | 6 | 4 | 5 | 51 |
| German | 750 | 250 | 2 | 6 | 3 | 11 | 61 |
| Glass | 161 | 53 | 6 | 9 | – | – | 9 |
| Heart | 226 | 76 | 2 | 6 | 3 | 4 | 22 |
| Hepatitis | 117 | 38 | 2 | 6 | 13 | – | 19 |
| Horse | 273 | 91 | 3 | 13 | 2 | 5 | 58 |
| Ionosphere | 264 | 87 | 2 | 33 | 1 | – | 34 |
| Iris | 113 | 37 | 3 | 4 | – | – | 4 |
| Labor | 43 | 14 | 2 | 8 | 3 | 5 | 29 |
| Liver | 259 | 86 | 2 | 6 | – | – | 2 |
| Lymphography | 111 | 37 | 4 | – | 9 | 6 | 38 |
| Pima | 576 | 192 | 2 | 8 | - | - | 8 |
| Promoters | 80 | 26 | 2 | – | – | 57 | 114 |
| Segment | 1733 | 577 | 7 | 19 | – | – | 19 |
| Sonar | 104 | 104 | 2 | 60 | – | – | 60 |
| Soybean | 513 | 170 | 19 | – | 16 | 19 | 82 |
| TicTacToe | 719 | 239 | 2 | – | – | 9 | 9 |
| Vehicle | 635 | 211 | 4 | 18 | – | – | 18 |
| Vote | 327 | 108 | 2 | – | 16 | – | 16 |
| Vowel | 528 | 462 | 11 | 10 | – | – | 10 |
| Zoo | 76 | 25 | 7 | 1 | 15 | – | 16 |

Table 5: Summary of data sets. The features of each data set can be C(continuous), B(binary) or N(nominal). The Inputs column shows the number of inputs of the network as it depends not only on the number of input variables but also on their type.

These data sets cover a wide variety of problems. There are problems with different numbers of available patterns, from 57 to 2310, different numbers of classes, from 2 to 19, different kinds of inputs, nominal, binary and continuous, and of different areas of application, from medical diagnosis to vowel recognition. Testing our model on this wide variety of problems can give us a clear idea of its performance. These are all the sets to which the method has been applied.

In order to test the efficiency of the proposed crossover in a classical artificial intelligence problem, we have used a RCGA to adjust the weight of each network within the ensemble. Our method considers each ensemble as a chromosome and applies a RCGA to optimize the weight of each network. The weight of each network of the ensemble is codified as a real number. The chromosome formed in this way is subject to CIXL2 crossover and non-uniform mutation. The parameters of CIXL2 are the same used in the rest of the paper, $n = 5$ and $1 - \alpha = 0.7$. The combination method used in the weighted sum of the outputs of the networks. Nevertheless, the same genetic algorithm could be used for weighting each network if a majority voting model is used.

The exact conditions of the experiments for each run of all the algorithms were the following:

- The ensemble was formed by 30 networks. Each network was trained separately using and standard back-propagation algorithm using the learning data.
- Once the 30 networks have been trained, the different methods for obtaining the weights were applied. So, all the methods use the same ensemble of networks on each run of the experiment. For the genetic algorithm, the fitness of each individual of the population is the classification accuracy over the learning set.
- After obtaining the vector of weights, the generalization error of each method is evaluated using the testing data.

Tables 6 and 7 show the results in terms of accurate classification for the 25 problems. The tables show the results using a RCGA with CIXL2, and the standard BEM and GEM methods. In order to compare the three methods we have performed a sign test over the win/draw/loss record of the three algorithms (Webb, 2000). These tests are shown in Table 8.

Table 8 shows the comparison statistics for the three models (Webb, 2000). For each model we show the win/draw/loss statistic, where the first value is the number of data sets for which $col < row$, the second is the number for which $col = row$, and the third is the number for which $col > row$. The second row shows the $p$-value of a two-tailed sign test on the win-loss record. The table shows that the genetic algorithm using CIXL2 is able to outperform the two standard algorithms BEM and GEM with a 10% confidence. On the other hand, there are no significant differences between BEM and GEM. This result is especially interesting because we have used a comprehensive set of problems from very different domains, different types of inputs, and different numbers of classes.

## 9. Conclusions and Future Work

In this paper we have proposed a crossover operator that allows the offspring to inherit features common to the best individuals of the population. The extraction of such common features is carried out by the determination of confidence intervals of the mean of the

| Problem | | Learning | | | | Test | | | | |
|---|---|---|---|---|---|---|---|---|---|---|
| | | Mean | St.Dev. | Best | Worst | Mean | St.Dev. | Best | Worst | |
| Anneal | CIXL2 | 0.9933 | 0.0046 | 0.9985 | 0.9777 | 0.9778 | 0.0090 | 0.9911 | 0.9420 | |
| | BEM | 0.9879 | 0.0054 | 0.9955 | 0.9733 | 0.9729 | 0.0091 | 0.9911 | 0.9464 | + |
| | GEM | 0.9915 | 0.0054 | 0.9985 | 0.9777 | 0.9780 | 0.0103 | 0.9911 | 0.9420 | - |
| Autos | CIXL2 | 0.8957 | 0.0233 | 0.9416 | 0.8506 | 0.7261 | 0.0577 | 0.8235 | 0.5882 | |
| | BEM | 0.8649 | 0.0211 | 0.9091 | 0.8312 | 0.7052 | 0.0586 | 0.8039 | 0.5686 | + |
| | GEM | 0.8740 | 0.0262 | 0.9351 | 0.8182 | 0.7033 | 0.0707 | 0.8039 | 0.5294 | + |
| Balance | CIXL2 | 0.9340 | 0.0067 | 0.9446 | 0.9232 | 0.9201 | 0.0118 | 0.9487 | 0.8910 | |
| | BEM | 0.9179 | 0.0068 | 0.9318 | 0.9019 | 0.9158 | 0.0111 | 0.9423 | 0.8910 | + |
| | GEM | 0.9148 | 0.0101 | 0.9318 | 0.8785 | 0.9158 | 0.0110 | 0.9359 | 0.8910 | + |
| Breast | CIXL2 | 0.8575 | 0.0195 | 0.8930 | 0.8047 | 0.6892 | 0.0322 | 0.7465 | 0.6338 | |
| | BEM | 0.8321 | 0.0287 | 0.8698 | 0.7395 | 0.6826 | 0.0375 | 0.7606 | 0.6056 | + |
| | GEM | 0.8274 | 0.0314 | 0.8791 | 0.7488 | 0.6817 | 0.0354 | 0.7324 | 0.6056 | + |
| Cancer | CIXL2 | 0.9723 | 0.0021 | 0.9771 | 0.9676 | 0.9799 | 0.0065 | 0.9885 | 0.9655 | |
| | BEM | 0.9678 | 0.0034 | 0.9733 | 0.9600 | 0.9793 | 0.0076 | 0.9943 | 0.9655 | + |
| | GEM | 0.9673 | 0.0034 | 0.9733 | 0.9581 | 0.9785 | 0.0084 | 0.9885 | 0.9598 | + |
| Card | CIXL2 | 0.9201 | 0.0087 | 0.9363 | 0.9054 | 0.8574 | 0.0153 | 0.8895 | 0.8256 | |
| | BEM | 0.9074 | 0.0088 | 0.9247 | 0.8880 | 0.8521 | 0.0212 | 0.8953 | 0.7965 | + |
| | GEM | 0.9049 | 0.0093 | 0.9208 | 0.8822 | 0.8533 | 0.0203 | 0.8953 | 0.7965 | + |
| German | CIXL2 | 0.8785 | 0.0080 | 0.8973 | 0.8653 | 0.7333 | 0.0184 | 0.7640 | 0.7000 | |
| | BEM | 0.8587 | 0.0090 | 0.8827 | 0.8440 | 0.7355 | 0.0141 | 0.7600 | 0.7040 | - |
| | GEM | 0.8642 | 0.0099 | 0.8827 | 0.8427 | 0.7377 | 0.0149 | 0.7680 | 0.7160 | - |
| Glass | CIXL2 | 0.8509 | 0.0225 | 0.9006 | 0.8075 | 0.6962 | 0.0365 | 0.7736 | 0.6038 | |
| | BEM | 0.8043 | 0.0246 | 0.8447 | 0.7578 | 0.6824 | 0.0424 | 0.7925 | 0.6038 | + |
| | GEM | 0.8246 | 0.0293 | 0.8820 | 0.7640 | 0.6855 | 0.0479 | 0.7736 | 0.6038 | + |
| Heart | CIXL2 | 0.9297 | 0.0216 | 0.9653 | 0.8861 | 0.8358 | 0.0271 | 0.8971 | 0.7794 | |
| | BEM | 0.9089 | 0.0214 | 0.9604 | 0.8663 | 0.8333 | 0.0263 | 0.8824 | 0.7794 | + |
| | GEM | 0.9182 | 0.0239 | 0.9554 | 0.8663 | 0.8279 | 0.0312 | 0.8971 | 0.7794 | + |
| Hepa. | CIXL2 | 0.9385 | 0.0224 | 0.9744 | 0.8718 | 0.8702 | 0.0372 | 0.9211 | 0.8158 | |
| | BEM | 0.9131 | 0.0253 | 0.9573 | 0.8462 | 0.8658 | 0.0319 | 0.9211 | 0.8158 | + |
| | GEM | 0.9179 | 0.0289 | 0.9744 | 0.8376 | 0.8711 | 0.0399 | 0.9474 | 0.7895 | - |
| Horse | CIXL2 | 0.8723 | 0.0174 | 0.9084 | 0.8315 | 0.7044 | 0.0313 | 0.7692 | 0.6264 | |
| | BEM | 0.8444 | 0.0194 | 0.8718 | 0.7949 | 0.7000 | 0.0301 | 0.7582 | 0.6374 | + |
| | GEM | 0.8485 | 0.0207 | 0.8864 | 0.8095 | 0.7004 | 0.0300 | 0.7802 | 0.6484 | + |
| Ionos. | CIXL2 | 0.9635 | 0.0164 | 0.9886 | 0.9356 | 0.8950 | 0.0225 | 0.9195 | 0.8276 | |
| | BEM | 0.9481 | 0.0171 | 0.9773 | 0.9167 | 0.8920 | 0.0206 | 0.9195 | 0.8276 | + |
| | GEM | 0.9554 | 0.0205 | 0.9886 | 0.9167 | 0.8958 | 0.0198 | 0.9310 | 0.8621 | - |
| Iris | CIXL2 | 1.0000 | 0.0000 | 1.0000 | 1.0000 | 1.0000 | 0.0000 | 1.0000 | 1.0000 | |
| | BEM | 1.0000 | 0.0000 | 1.0000 | 1.0000 | 1.0000 | 0.0000 | 1.0000 | 1.0000 | = |
| | GEM | 1.0000 | 0.0000 | 1.0000 | 1.0000 | 1.0000 | 0.0000 | 1.0000 | 1.0000 | = |

Table 6: Ensemble results using real-coded genetic algorithm (CIXL2), basic ensemble method (BEM), and generalized ensemble method (GEM). For each problem we have marked whichever CIXL2 is better (+), equal, (=), or worse (-) than BEM/GEM.

| Problem | | Learning | | | | Test | | | | |
|---|---|---|---|---|---|---|---|---|---|---|
| | | Mean | St.Dev. | Best | Worst | Mean | St.Dev. | Best | Worst | |
| Labor | CIXL2 | 0.9651 | 0.0257 | 1.0000 | 0.8837 | 0.8857 | 0.0550 | 1.0000 | 0.7857 | |
| | BEM | 0.9488 | 0.0283 | 0.9767 | 0.8837 | 0.8833 | 0.0663 | 1.0000 | 0.7143 | + |
| | GEM | 0.9527 | 0.0270 | 0.9767 | 0.8837 | 0.8833 | 0.0689 | 1.0000 | 0.7143 | + |
| Liver | CIXL2 | 0.8126 | 0.0175 | 0.8494 | 0.7761 | 0.6992 | 0.0276 | 0.7442 | 0.6512 | |
| | BEM | 0.7799 | 0.0176 | 0.8108 | 0.7336 | 0.6950 | 0.0253 | 0.7442 | 0.6395 | + |
| | GEM | 0.7744 | 0.0198 | 0.8108 | 0.7336 | 0.6826 | 0.0337 | 0.7442 | 0.6047 | + |
| Lymph | CIXL2 | 0.9456 | 0.0208 | 0.9730 | 0.8919 | 0.7847 | 0.0538 | 0.8649 | 0.6486 | |
| | BEM | 0.9318 | 0.0242 | 0.9640 | 0.8739 | 0.7775 | 0.0539 | 0.8649 | 0.6486 | + |
| | GEM | 0.9306 | 0.0254 | 0.9730 | 0.8559 | 0.7784 | 0.0504 | 0.8378 | 0.6486 | + |
| Pima | CIXL2 | 0.7982 | 0.0073 | 0.8194 | 0.7830 | 0.7811 | 0.0209 | 0.8177 | 0.7292 | |
| | BEM | 0.7782 | 0.0079 | 0.7934 | 0.7535 | 0.7885 | 0.0199 | 0.8177 | 0.7448 | − |
| | GEM | 0.7752 | 0.0089 | 0.7882 | 0.7431 | 0.7793 | 0.0222 | 0.8281 | 0.7292 | + |
| Promot. | CIXL2 | 0.9496 | 0.0304 | 1.0000 | 0.8875 | 0.8244 | 0.0726 | 1.0000 | 0.7308 | |
| | BEM | 0.9300 | 0.0357 | 0.9875 | 0.8500 | 0.8269 | 0.0612 | 0.9231 | 0.7308 | − |
| | GEM | 0.9263 | 0.0319 | 0.9875 | 0.8625 | 0.8218 | 0.0711 | 0.9615 | 0.6923 | + |
| Segment | CIXL2 | 0.9502 | 0.0030 | 0.9544 | 0.9446 | 0.9259 | 0.0057 | 0.9376 | 0.9151 | |
| | BEM | 0.9339 | 0.0042 | 0.9411 | 0.9256 | 0.9183 | 0.0054 | 0.9341 | 0.9081 | + |
| | GEM | 0.9423 | 0.0044 | 0.9521 | 0.9319 | 0.9236 | 0.0061 | 0.9359 | 0.9116 | + |
| Sonar | CIXL2 | 0.9074 | 0.0236 | 0.9519 | 0.8654 | 0.7849 | 0.0286 | 0.8462 | 0.7404 | |
| | BEM | 0.8859 | 0.0266 | 0.9423 | 0.8269 | 0.7865 | 0.0286 | 0.8365 | 0.7212 | − |
| | GEM | 0.8907 | 0.0277 | 0.9519 | 0.8365 | 0.7853 | 0.0266 | 0.8462 | 0.7404 | − |
| Soybean | CIXL2 | 0.9758 | 0.0114 | 0.9903 | 0.9454 | 0.9057 | 0.0165 | 0.9353 | 0.8706 | |
| | BEM | 0.9602 | 0.0130 | 0.9805 | 0.9240 | 0.9039 | 0.0182 | 0.9353 | 0.8647 | + |
| | GEM | 0.9691 | 0.0157 | 0.9883 | 0.9376 | 0.9067 | 0.0187 | 0.9353 | 0.8706 | − |
| TicTacToe | CIXL2 | 0.9913 | 0.0027 | 0.9972 | 0.9847 | 0.9794 | 0.0024 | 0.9874 | 0.9749 | |
| | BEM | 0.9868 | 0.0020 | 0.9917 | 0.9847 | 0.9791 | 0.0000 | 0.9791 | 0.9791 | + |
| | GEM | 0.9876 | 0.0024 | 0.9930 | 0.9847 | 0.9792 | 0.0008 | 0.9833 | 0.9791 | + |
| Vote | CIXL2 | 0.9832 | 0.0055 | 0.9939 | 0.9725 | 0.9278 | 0.0110 | 0.9537 | 0.8889 | |
| | BEM | 0.9793 | 0.0060 | 0.9908 | 0.9664 | 0.9284 | 0.0068 | 0.9444 | 0.9167 | − |
| | GEM | 0.9801 | 0.0062 | 0.9908 | 0.9664 | 0.9262 | 0.0107 | 0.9444 | 0.8981 | + |
| Vowel | CIXL2 | 0.9146 | 0.0148 | 0.9432 | 0.8845 | 0.4925 | 0.0293 | 0.5606 | 0.4459 | |
| | BEM | 0.8733 | 0.0179 | 0.9015 | 0.8371 | 0.4913 | 0.0331 | 0.5584 | 0.4264 | + |
| | GEM | 0.9157 | 0.0129 | 0.9394 | 0.8845 | 0.4973 | 0.0342 | 0.5541 | 0.4221 | − |
| Zoo | CIXL2 | 0.9807 | 0.0175 | 1.0000 | 0.9211 | 0.9360 | 0.0290 | 0.9600 | 0.8800 | |
| | BEM | 0.9671 | 0.0215 | 1.0000 | 0.9079 | 0.9307 | 0.0392 | 0.9600 | 0.8400 | + |
| | GEM | 0.9750 | 0.0203 | 1.0000 | 0.9211 | 0.9307 | 0.0347 | 0.9600 | 0.8400 | + |

Table 7: Ensemble results using real-coded genetic algorithm (CIXL2), basic ensemble method (BEM), and generalized ensemble method (GEM). For each problem we have marked whichever CIXL2 is better (+), equal, (=), or worse (-) than BEM/GEM.

| Algorithm | BEM | GEM | |
|---|---|---|---|
| CIXL2 | 19/1/5 | 17/1/7 | win/draw/loss |
| | 0.0066 | 0.0639 | $p$-value |
| BEM | | 9/4/12 | win/draw/loss |
| | | 0.6636 | $p$-value |

Table 8: Comparison of the three methods. Win/draw/loss record of the algorithms against each other and $p$-value of the sign test.

best individuals of the population. From these confidence intervals, CIXL2 creates three virtual parents that are used to implement a directed search towards the region of the fittest individuals. The amplitude and speed of the search is determined by the number of best individuals selected and the confidence coefficient.

The study carried out in order to obtain the best parameters for CIXL2 concludes that the value of $n = 5$ best individuals is suitable to obtain the localization estimator to guide the search in most of the problems tested. However, in very difficult problems, it would be advisable to have a larger value of $n$ to avoid the premature convergence of the evolutionary process. The confident coefficient, $1 - \alpha$, is responsible, together with the dispersion of the best individuals, for the modulation of the wideness of the confidence interval centered on the localization estimator. The study results in the best value of $1 - \alpha = 0.70$. This pair of values has an acceptable performance for all problems, although there is not an optimum pair of values for all problems.

The comparative analysis of the crossover operators shows that CIXL2 is a good alternative to widely used crossovers such as $BLX_\alpha$ for unimodal function such as $f_{Sph}$, $f_{SchDS}$, and $f_{Ros}$. Noteworthy is the performance of CIXL2 in the two non-separable functions, $f_{SchDS}$ and $f_{Ros}$, where the other crossovers have a disparate behavior.

If in unimodal functions the strategy of extracting the statistical features of localization and dispersion of the best individuals is a guarantee of good performance, the case for multimodal functions is quite different, and the performance of the algorithm is not assured a priori. Nevertheless, the results obtained for this kind of functions show that CIXL2 is always one of the best performing operators. For instance, in functions of a high complexity such as $f_{Ack}$ — multimodal, non-separable and regular — and $f_{Fle}$ — multimodal, non-separable and irregular — CIXL2 obtains the best results. This behavior reveals that the determination of the region of the best individuals by means of confidence intervals provides a robust methodology that, applied to crossover operator, shows an interesting performance even in very difficult functions. In summary, we can affirm that this paper proves that CIXL2 is a promising alternative to bear in mind, when we must choose which crossover to use in a real-coded genetic algorithm.

EDAs have shown very good performance for unimodal and separable functions, $f_{Sph}$, and for functions whose optima are regularly distributed, $f_{Ack}$ and $f_{Gri}$. The performance of EDAs decreases in multimodal, $f_{Ras}$ and $f_{Sch}$, and epistatic functions, $f_{SchDS}$ and $f_{Ros}$. On the other hand, CIXL2 is less sensitive to the type of function. The main reason for this behavior may be found in the fact that CIXL2 uses the distribution information obtained from the best individuals of the population differently. CIXL2 creates three virtual parents from this distribution, but if the virtual parents have worse fitness than the individual which mates, the offspring is not generated near these virtual parents. In this way, CIXL2 prevents a shifting of the population to the confidence interval if the improvement of the performance is not significant.

The applicability of the proposed crossover to a problem of artificial neural network ensembles shows how this model can be used for solving standard artificial intelligence problems. RCGAs with CIXL2 can also be used in other aspects of ensemble design, such as, selection of a subset of networks, and sampling of the training set of each network.

These promising results motivate the beginning of a new line of research geared to the study of the distribution of the best individuals taking into account the kind of problem at hand. We aim to propose new techniques of selection of individuals to be considered for obtaining the confidence interval in a more reliable way. In multimodal, irregular, or with many chaotically scattered optima functions the difficulty of obtaining the distributions of the best individuals is enormous. In these kind of functions it would be interesting to perform a cluster analysis of the selected best individuals and to obtain a confidence interval for every cluster. This idea would allow the implementation of a multi-directional crossover towards different promising regions.

On the other hand, it is likely that as the evolutive process progresses the distribution of the best individuals changes. In such a case, it would be advisable to perform, at regular intervals, statistical tests to determine the distribution that best reflects the features of the best individuals on the population.

Alternatively, we are considering the construction of non-parametric confidence intervals. In this way, we need more robust estimators of the parameters of localization and dispersion of the genes of the best individuals. We have performed some preliminary studies using the median and different measures of dispersion and the results are quite encouraging.

Another research line currently open is the study of the application of CIXL2 to problems of optimization with restrictions, especially in the presence of non-linearity, where the generation of individuals in the feasible region is a big issue. The orientation of the search based on the identification of the region of the best individuals that is implemented by CIXL2 could favor the generation of feasible individuals. This feature would be an interesting advantage with respect to other crossover operators.

## Acknowledgments

The authors would like to acknowledge R. Moya-Sánchez for her helping in the final version of this paper.

This work has been financed in part by the project TIC2002-04036-C05-02 of the Spanish Inter-Ministerial Commission of Science and Technology (CICYT) and FEDER funds.

## Appendix A. Results of the Statistical Study

| Function | $\alpha^*$ C | $\alpha^*$ B | $\alpha^*$ CB | $\mathbf{R}^2$ | $\alpha^*$ T. Levene |
|---|---|---|---|---|---|
| $f_{Sph}$ | 1.000 | 1.000 | — | — | 0.000 |
| $f_{SchDS}$ | 0.000 | 0.000 | 0.000 | 0.601 | 0.000 |
| $f_{Ros}$ | 0.005 | 0.000 | 0.006 | 0.526 | 0.000 |
| $f_{Ras}$ | 0.000 | 0.000 | 0.000 | 0.617 | 0.000 |
| $f_{Sch}$ | 0.000 | 0.000 | 0.000 | 0.805 | 0.000 |
| $f_{Ack}$ | 0.095 | 0.000 | 0.019 | 0.083 | 0.000 |
| $f_{Gri}$ | 0.149 | 0.001 | — | 0.040 | 0.000 |
| $f_{Fle}$ | 0.410 | 0.000 | — | 0.054 | 0.003 |
| $f_{Lan}$ | 0.040 | 0.000 | 0.024 | 0.159 | 0.000 |

Table 9: Significant levels, $\alpha^*$, of each term of the linear model, determination coefficient $R^2$, and value of Levene test of the statistical analysis of CIXL2 parameters.

| Function | $\alpha^*$ Crossover | $\mathbf{R}^2$ | $\alpha^*$ Levene test |
|---|---|---|---|
| $f_{Sph}$ | 0.000 | 0.779 | 0.000 |
| $f_{SchDS}$ | 0.000 | 0.786 | 0.000 |
| $f_{Ros}$ | 0.573 | 0.024 | 0.000 |
| $f_{Ras}$ | 0.000 | 0.971 | 0.000 |
| $f_{Sch}$ | 0.000 | 0.987 | 0.000 |
| $f_{Ack}$ | 0.000 | 0.884 | 0.000 |
| $f_{Gri}$ | 0.000 | 0.421 | 0.000 |
| $f_{Fle}$ | 0.000 | 0.137 | 0.091 |
| $f_{Lan}$ | 0.000 | 0.486 | 0.000 |

Table 10: Significance level of the crossover operator and determination coefficient $R^2$ of the linear model, and value of Levene test of the comparative study of the crossovers.

| Function | $\alpha^*$ EA | $\mathbf{R}^2$ | $\alpha^*$ Levene test |
|---|---|---|---|
| $f_{SchDS}$ | 0.000 | 0.955 | 0.000 |
| $f_{Ros}$ | 0.000 | 0.778 | 0.000 |
| $f_{Ras}$ | 0.000 | 0.992 | 0.000 |
| $f_{Sch}$ | 0.000 | 0.999 | 0.000 |
| $f_{Ack}$ | 1.000 | 0.641 | 1.000 |
| $f_{Gri}$ | 0.000 | 0.455 | 0.000 |
| $f_{Fle}$ | 0.001 | 0.150 | 0.000 |
| $f_{Lan}$ | 0.027 | 0.079 | 0.000 |

Table 11: Significance level of the evolutionary algorithms and determination coefficient $R^2$ of the linear model, and value of Levene test of the comparative study betwen CIXL2 and EDAs.

| I | J | $\mu_I - \mu_J$ | $\alpha^*$ | $\mu_I - \mu_J$ | $\alpha^*$ | $\mu_I - \mu_J$ | $\alpha^*$ |
|---|---|---|---|---|---|---|---|
| | | $f_{Sph}$ | | $f_{SchDS}$ | | $f_{Ros}$ | |
| 5 | 10 | -2.683e-15 | 0.000 | -2.540e-02 | 0.000 | -9.433e-01 | 0.000 |
| | 30 | -4.144e-11 | 0.000 | -1.899e-01 | 0.000 | -1.486e+00 | 0.000 |
| | 60 | -1.836e-07 | 0.000 | -2.371e-01 | 0.000 | -1.058e+00 | 0.000 |
| | 90 | -5.554e-08 | 0.000 | -1.004e+00 | 0.000 | -8.375e-01 | 0.000 |
| 10 | 5 | 2.683e-15 | 0.000 | 2.540e-02 | 0.000 | 9.433e-01 | 0.000 |
| | 30 | -4.144e-11 | 0.000 | -1.645e-01 | 0.000 | -5.425e-01 | 0.000 |
| | 60 | -1.836e-07 | 0.000 | -2.117e-01 | 0.000 | -1.142e-01 | 0.025 |
| | 90 | -5.554e-08 | 0.000 | -9.785e-01 | 0.000 | 1.058e-01 | 0.014 |
| 30 | 5 | 4.144e-11 | 0.000 | 1.899e-01 | 0.000 | 1.486e+00 | 0.000 |
| | 10 | 4.144e-11 | 0.000 | 1.645e-01 | 0.000 | 5.425e-01 | 0.000 |
| | 60 | -1.835e-07 | 0.000 | -4.720e-02 | 0.572 | 4.283e-01 | 0.000 |
| | 90 | -5.549e-08 | 0.000 | -8.140e-01 | 0.000 | 6.483e-01 | 0.000 |
| 60 | 5 | 1.836e-07 | 0.000 | 2.371e-01 | 0.000 | 1.058e+00 | 0.000 |
| | 10 | 1.836e-07 | 0.000 | 2.117e-01 | 0.000 | 1.142e-01 | 0.025 |
| | 30 | 1.835e-07 | 0.000 | 4.720e-02 | 0.572 | -4.283e-01 | 0.000 |
| | 90 | 1.281e-07 | 0.003 | -7.668e-01 | 0.000 | 2.200e-01 | 0.000 |
| 90 | 5 | 5.554e-08 | 0.000 | 1.004e+00 | 0.000 | 8.375e-01 | 0.000 |
| | 10 | 5.554e-08 | 0.000 | 9.785e-01 | 0.000 | -1.058e-01 | 0.014 |
| | 30 | 5.549e-08 | 0.000 | 8.140e-01 | 0.000 | -6.483e-01 | 0.000 |
| | 60 | -1.281e-07 | 0.003 | 7.668e-01 | 0.000 | -2.200e-01 | 0.000 |
| **Ranking** | | $\mu_{60} > \mu_{90} > \mu_{30} > \mu_{10} > \mu_5$ | | $\mu_{90} > \mu_{30} > \mu_{10} > \mu_5$<br>$\mu_{60} > \mu_5$ | | $\mu_{30} > \mu_{60} > \mu_{10} > \mu_{90} > \mu_5$ | |
| | | $f_{Ras}$ | | $f_{Sch}$ | | $f_{Ack}$ | |
| 5 | 10 | -5.79e+00 | 0.000 | -2.691e+02 | 0.082 | -1.063e-07 | 0.000 |
| | 30 | -6.72e+00 | 0.000 | -7.338e+02 | 0.000 | -2.384e-05 | 0.000 |
| | 60 | -1.01e+01 | 0.000 | -9.559e+02 | 0.000 | -1.508e-03 | 0.000 |
| | 90 | -1.51e+01 | 0.000 | -1.148e+03 | 0.000 | -6.769e-02 | 0.216 |
| 10 | 5 | 5.79e+00 | 0.000 | 2.691e+02 | 0.082 | 1.063e-07 | 0.000 |
| | 30 | -9.31e-01 | 0.807 | -4.647e+02 | 0.000 | -2.373e-05 | 0.000 |
| | 60 | -4.30e+00 | 0.000 | -6.868e+02 | 0.000 | -1.508e-03 | 0.000 |
| | 90 | -9.32e+00 | 0.000 | -8.786e+02 | 0.000 | -6.769e-02 | 0.216 |
| 30 | 5 | 6.72e+00 | 0.000 | 7.338e+02 | 0.000 | 2.384e-05 | 0.000 |
| | 10 | 9.31e-01 | 0.807 | 4.647e+02 | 0.000 | 2.373e-05 | 0.000 |
| | 60 | -3.37e+00 | 0.000 | -2.221e+02 | 0.000 | -1.484e-03 | 0.000 |
| | 90 | -8.39e+00 | 0.000 | -4.139e+02 | 0.000 | -6.767e-02 | 0.216 |
| 60 | 5 | 1.01e+01 | 0.000 | 9.559e+02 | 0.000 | 1.508e-03 | 0.000 |
| | 10 | 4.30e+00 | 0.000 | 6.868e+02 | 0.000 | 1.508e-03 | 0.000 |
| | 30 | 3.37e+00 | 0.000 | 2.221e+02 | 0.000 | 1.484e-03 | 0.000 |
| | 90 | -5.02e+00 | 0.000 | -1.918e+02 | 0.000 | -6.619e-02 | 0.242 |
| 90 | 5 | 1.51e+01 | 0.000 | 1.148e+03 | 0.000 | 6.769e-02 | 0.216 |
| | 10 | 9.32e+00 | 0.000 | 8.786e+02 | 0.000 | 6.769e-02 | 0.216 |
| | 30 | 8.39e+00 | 0.000 | 4.139e+02 | 0.000 | 6.767e-02 | 0.216 |
| | 60 | 5.02e+00 | 0.000 | 1.918e+02 | 0.000 | 6.619e-02 | 0.242 |
| **Ranking** | | $\mu_{90} > \mu_{60} > \mu_{10} > \mu_5$<br>$\mu_{30} > \mu_5$ | | $\mu_{90} > \mu_{60} > \mu_{30} > \mu_5$<br>$\mu_{10} \geq \mu_5$ | | $\mu_{60} > \mu_{30} > \mu_{10} > \mu_5$<br>$\mu_{90} \geq \mu_5$ | |
| | | $f_{Gri}$ | | $f_{Fle}$ | | $f_{Lan}$ | |
| 5 | 10 | -7.207E-03 | 0.174 | -2.776e+03 | 0.885 | -1.354e-02 | 0.998 |
| | 30 | -3.896E-03 | 0.864 | -7.968e+03 | 0.004 | -5.881e-02 | 0.009 |
| | 60 | 2.329E-03 | 1.000 | -7.342e+03 | 0.008 | -8.794e-02 | 0.000 |
| | 90 | 8.649E-03 | 0.001 | -1.268e+04 | 0.000 | -1.142e-01 | 0.000 |
| 10 | 5 | 7.207E-03 | 0.174 | 2.776e+03 | 0.885 | 1.354e-02 | 0.998 |
| | 30 | 3.311E-03 | 0.983 | -5.192e+03 | 0.234 | -4.527e-02 | 0.082 |
| | 60 | 9.535E-03 | 0.533 | -4.566e+03 | 0.378 | -7.440e-02 | 0.000 |
| | 90 | 1.586E-02 | 0.000 | -9.899e+03 | 0.006 | -1.007e-01 | 0.000 |
| 30 | 5 | 3.896E-03 | 0.864 | 7.968e+03 | 0.004 | 5.881e-02 | 0.009 |
| | 10 | -3.311E-03 | 0.983 | 5.192e+03 | 0.234 | 4.527e-02 | 0.082 |
| | 60 | 6.225E-03 | 0.930 | 6.254e+02 | 1.000 | -2.913e-02 | 0.354 |
| | 90 | 1.254E-02 | 0.000 | -4.707e+03 | 0.678 | -5.540e-02 | 0.000 |
| 60 | 5 | -2.329E-03 | 1.000 | 7.342e+03 | 0.008 | 8.794e-02 | 0.000 |
| | 10 | -9.535E-03 | 0.533 | 4.566e+03 | 0.378 | 7.440e-02 | 0.000 |
| | 30 | -6.225E-03 | 0.930 | -6.254e+02 | 1.000 | 2.913e-02 | 0.354 |
| | 90 | 6.320E-03 | 0.884 | -5.333e+03 | 0.491 | -2.627e-02 | 0.247 |
| 90 | 5 | -8.649E-03 | 0.001 | 1.268e+04 | 0.000 | 1.142e-01 | 0.000 |
| | 10 | -1.586E-02 | 0.000 | 9.899e+03 | 0.006 | 1.007e-01 | 0.000 |
| | 30 | -1.254E-02 | 0.000 | 4.707e+03 | 0.678 | 5.540e-02 | 0.000 |
| | 60 | -6.320E-03 | 0.884 | 5.333e+03 | 0.491 | 2.627e-02 | 0.247 |
| **Ranking** | | $\mu_{60} \geq \mu_{90}$ $\mu_5 > \mu_{90}$<br>$\mu_{10} > \mu_{90}$ $\mu_{30} > \mu_{90}$ | | $\mu_{10} \geq \mu_5$ $\mu_{30} > \mu_5$<br>$\mu_{60} > \mu_5$ $\mu_{90} > \mu_5$ | | $\mu_{10} \geq \mu_5$ $\mu_{30} > \mu_5$<br>$\mu_{60} > \mu_5$ $\mu_{90} > \mu_5$ | |

Table 12: Results for all the functions of the multiple comparison test and the ranking obtained depending on the number of best individuals $n$.

| I | J | $\mu_{\mathbf{I}} - \mu_{\mathbf{J}}$ | $\alpha^*$ | $\mu_{\mathbf{I}} - \mu_{\mathbf{J}}$ | $\alpha^*$ | $\mu_{\mathbf{I}} - \mu_{\mathbf{J}}$ | $\alpha^*$ |
|---|---|---|---|---|---|---|---|
| | | $\mathbf{f_{Sph}}$ | | $\mathbf{f_{SchDS}}$ | | $\mathbf{f_{Ros}}$ | |
| 0.70 | 0.90 | -1.361e-08 | 0.000 | -3.985e-01 | 0.000 | -1.360e-01 | 0.281 |
| | 0.95 | -4.394e-08 | 0.000 | -3.783e-02 | 0.967 | -1.693e-01 | 0.131 |
| | 0.99 | -1.302e-07 | 0.000 | 8.165e-02 | 0.114 | -1.813e-01 | 0.310 |
| 0.90 | 0.70 | 1.361e-08 | 0.000 | 3.985e-01 | 0.000 | 1.360e-01 | 0.281 |
| | 0.95 | -3.033e-08 | 0.000 | 3.607e-01 | 0.001 | -3.333e-02 | 0.995 |
| | 0.99 | -1.166e-07 | 0.000 | 4.802e-01 | 0.000 | -4.533e-02 | 0.996 |
| 0.95 | 0.70 | 4.394e-08 | 0.000 | 3.783e-02 | 0.967 | 1.693e-01 | 0.131 |
| | 0.90 | 3.033e-08 | 0.000 | -3.607e-01 | 0.001 | 3.333e-02 | 0.995 |
| | 0.99 | -8.628e-08 | 0.019 | 1.195e-01 | 0.013 | -1.200e-02 | 1.000 |
| 0.99 | 0.70 | 1.302e-07 | 0.000 | -8.165e-02 | 0.114 | 1.813e-01 | 0.310 |
| | 0.90 | 1.166e-07 | 0.000 | -4.802e-01 | 0.000 | 4.533e-02 | 0.996 |
| | 0.95 | 8.628e-08 | 0.019 | -1.195e-01 | 0.013 | 1.200e-02 | 1.000 |
| **Ranking** | | $\mu_{0.99} > \mu_{0.95} > \mu_{0.90} > \mu_{0.70}$ | | $\mu_{0.90} > \mu_{0.95} > \mu_{0.99}$<br>$\mu_{0.70} \geq \mu_{0.99}$ | | $\mu_{0.99} \geq \mu_{0.95} \geq \mu_{0.90} \geq \mu_{0.70}$ | |
| | | $\mathbf{f_{Ras}}$ | | $\mathbf{f_{Sch}}$ | | $\mathbf{f_{Ack}}$ | |
| 0.70 | 0.90 | -4.23e+00 | 0.000 | 1.198e+02 | 0.714 | -2.471e-04 | 0.000 |
| | 0.95 | -3.59e+00 | 0.000 | 8.247e+01 | 0.919 | -1.944e-02 | 0.617 |
| | 0.99 | -5.56e+00 | 0.000 | -3.008e+02 | 0.001 | -3.541e-02 | 0.382 |
| 0.90 | 0.70 | 4.23e+00 | 0.000 | -1.198e+02 | 0.714 | 2.471e-04 | 0.000 |
| | 0.95 | 6.40e-01 | 0.966 | -3.736e+01 | 0.997 | -1.919e-02 | 0.631 |
| | 0.99 | -1.33e+00 | 0.551 | -4.206e+02 | 0.000 | -3.516e-02 | 0.390 |
| 0.95 | 0.70 | 3.59e+00 | 0.000 | -8.247e+01 | 0.919 | 1.944e-02 | 0.617 |
| | 0.90 | -6.40e-01 | 0.966 | 3.736e+01 | 0.997 | 1.919e-02 | 0.631 |
| | 0.99 | -1.97e+00 | 0.044 | -3.833e+02 | 0.000 | -1.597e-02 | 0.985 |
| 0.99 | 0.70 | 5.56e+00 | 0.000 | 3.008e+02 | 0.001 | 3.541e-02 | 0.382 |
| | 0.90 | 1.33e+00 | 0.551 | 4.206e+02 | 0.000 | 3.516e-02 | 0.390 |
| | 0.95 | 1.97e+00 | 0.044 | 3.833e+02 | 0.000 | 1.597e-02 | 0.985 |
| **Ranking** | | $\mu_{0.99} > \mu_{0.95} > \mu_{0.70}$<br>$\mu_{0.90} > \mu_{0.70}$ | | $\mu_{0.70} \geq \mu_{0.95} \geq \mu_{0.90}$<br>$\mu_{0.99} > \mu_{0.90}$ | | $\mu_{0.99} \geq \mu_{0.95} \geq \mu_{0.70}$<br>$\mu_{0.90} > \mu_{0.70}$ | |
| | | $\mathbf{f_{Gri}}$ | | $\mathbf{f_{Fle}}$ | | $\mathbf{f_{Lan}}$ | |
| 0.70 | 0.90 | -7.196E-03 | 0.395 | -2.986e+03 | 0.717 | 6.105e-03 | 0.998 |
| | 0.95 | -2.027E-03 | 0.945 | -3.241e+03 | 0.635 | 2.867e-02 | 0.272 |
| | 0.99 | -5.667E-03 | 0.155 | -3.079e+03 | 0.644 | 3.309e-02 | 0.133 |
| 0.90 | 0.70 | 7.196E-03 | 0.395 | 2.986e+03 | 0.717 | -6.105e-03 | 0.998 |
| | 0.95 | 5.168E-03 | 0.791 | -2.547e+02 | 1.000 | 2.257e-02 | 0.585 |
| | 0.99 | 1.529E-03 | 1.000 | -9.255e+01 | 1.000 | 2.698e-02 | 0.363 |
| 0.95 | 0.70 | 2.027E-03 | 0.945 | 3.241e+03 | 0.635 | -2.867e-02 | 0.272 |
| | 0.90 | -5.168E-03 | 0.791 | 2.547e+02 | 1.000 | -2.257e-02 | 0.585 |
| | 0.99 | -3.640E-03 | 0.747 | 1.622e+02 | 1.000 | 4.415e-03 | 1.000 |
| 0.99 | 0.70 | 5.667E-03 | 0.155 | 3.079e+03 | 0.644 | -3.309e-02 | 0.133 |
| | 0.90 | -1.529E-03 | 1.000 | 9.255e+01 | 1.000 | -2.698e-02 | 0.363 |
| | 0.95 | 3.640E-03 | 0.747 | -1.622e+02 | 1.000 | -4.415e-03 | 1.000 |
| **Ranking** | | $\mu_{0.90} \geq \mu_{0.99} \geq \mu_{0.95} \geq \mu_{0.70}$ | | $\mu_{0.95} \geq \mu_{0.99} \geq \mu_{0.90} \geq \mu_{0.70}$ | | $\mu_{0.70} \geq \mu_{0.90} \geq \mu_{0.95} \geq \mu_{0.99}$ | |

Table 13: Results for all the functions of the multiple comparison test and the ranking obtained depending on the confidence coefficient $1 - \alpha$.

| Crossover | | $f_{\mathbf{Sph}}$ | | $f_{\mathbf{SchDS}}$ | | $f_{\mathbf{Ros}}$ | |
|---|---|---|---|---|---|---|---|
| **I** | **J** | $\mu_{\mathbf{I}} - \mu_{\mathbf{J}}$ | $\alpha^*$ | $\mu_{\mathbf{I}} - \mu_{\mathbf{J}}$ | $\alpha^*$ | $\mu_{\mathbf{I}} - \mu_{\mathbf{J}}$ | $\alpha^*$ |
| CIXL2 | BLX(0.3) | 3.109e-16 | 0.000 | -1.583e-02 | 0.000 | -4.283e+00 | 0.997 |
| | BLX(0.5) | 1.628e-16 | 0.212 | -7.337e-03 | 0.028 | -6.667e+00 | 0.933 |
| | SBX(2) | -1.644e-12 | 0.000 | -2.014e-01 | 0.000 | -2.809e+00 | 0.958 |
| | SBX(5) | -4.873e-12 | 0.000 | -3.913e-01 | 0.000 | -6.165e+00 | 0.944 |
| | Fuzzy | -2.102e-15 | 0.000 | -3.968e+01 | 0.000 | -2.487e+00 | 1.000 |
| | Logical | -3.689e-13 | 0.000 | -1.098e+01 | 0.000 | -2.092e+00 | 0.000 |
| | UNDX | -2.910e-05 | 0.000 | -2.080e+01 | 0.000 | -3.460e+00 | 0.000 |
| BLX(0.3) | CIXL2 | -3.109e-16 | 0.000 | 1.583e-02 | 0.000 | 4.283e+00 | 0.997 |
| | BLX(0.5) | -1.480e-16 | 0.074 | 8.495e-03 | 0.357 | -2.384e+00 | 1.000 |
| | SBX(2) | -1.644e-12 | 0.000 | -1.855e-01 | 0.000 | 1.473e+00 | 1.000 |
| | SBX(5) | -4.873e-12 | 0.000 | -3.755e-01 | 0.000 | -1.882e+00 | 1.000 |
| | Fuzzy | -2.413e-15 | 0.000 | -3.966e+01 | 0.000 | 1.796e+00 | 1.000 |
| | Logical | -3.692e-13 | 0.000 | -1.097e+01 | 0.000 | 2.191e+00 | 1.000 |
| | UNDX | -2.910e-05 | 0.000 | -2.078e+01 | 0.000 | 8.225e-01 | 1.000 |
| BLX(0.5) | CIXL2 | -1.628e-16 | 0.212 | 7.337e-03 | 0.028 | 6.667e+00 | 0.933 |
| | BLX(0.3) | 1.480e-16 | 0.074 | -8.495e-03 | 0.357 | 2.384e+00 | 1.000 |
| | SBX(2) | -1.644e-12 | 0.000 | -1.940e-01 | 0.000 | 3.857e+00 | 1.000 |
| | SBX(5) | -4.873e-12 | 0.000 | -3.840e-01 | 0.000 | 5.019e-01 | 1.000 |
| | Fuzzy | -2.265e-15 | 0.000 | -3.967e+01 | 0.000 | 4.179e+00 | 1.000 |
| | Logical | -3.690e-13 | 0.000 | -1.098e+01 | 0.000 | 4.575e+00 | 1.000 |
| | UNDX | -2.910e-05 | 0.000 | -2.079e+01 | 0.000 | 3.206e+00 | 1.000 |
| SBX(2) | CIXL2 | 1.644e-12 | 0.000 | 2.014e-01 | 0.000 | 2.809e+00 | 0.958 |
| | BLX(0.3) | 1.644e-12 | 0.000 | 1.855e-01 | 0.000 | -1.473e+00 | 1.000 |
| | BLX(0.5) | 1.644e-12 | 0.000 | 1.940e-01 | 0.000 | -3.857e+00 | 1.000 |
| | SBX(5) | -3.229e-12 | 0.000 | -1.900e-01 | 0.115 | -3.355e+00 | 1.000 |
| | Fuzzy | 1.642e-12 | 0.000 | -3.948e+01 | 0.000 | 3.222e-01 | 1.000 |
| | Logical | 1.275e-12 | 0.000 | -1.078e+01 | 0.000 | 7.179e-01 | 1.000 |
| | UNDX | -2.910e-05 | 0.000 | -2.060e+01 | 0.000 | -6.508e-01 | 1.000 |
| SBX(5) | CIXL2 | 4.873e-12 | 0.000 | 3.913e-01 | 0.000 | 6.165e+00 | 0.944 |
| | BLX(0.3) | 4.873e-12 | 0.000 | 3.755e-01 | 0.000 | 1.882e+00 | 1.000 |
| | BLX(0.5) | 4.873e-12 | 0.000 | 3.840e-01 | 0.000 | -5.019e-01 | 1.000 |
| | SBX(2) | 3.229e-12 | 0.000 | 1.900e-01 | 0.115 | 3.355e+00 | 1.000 |
| | Fuzzy | 4.871e-12 | 0.000 | -3.929e+01 | 0.000 | 3.678e+00 | 1.000 |
| | Logical | 4.504e-12 | 0.000 | -1.059e+01 | 0.000 | 4.073e+00 | 1.000 |
| | UNDX | -2.910e-05 | 0.000 | -2.041e+01 | 0.000 | 2.705e+00 | 1.000 |
| Fuzzy | CIXL2 | 2.102e-15 | 0.000 | 3.968e+01 | 0.000 | 2.487e+00 | 1.000 |
| | BLX(0.3) | 2.413e-15 | 0.000 | 3.966e+01 | 0.000 | -1.796e+00 | 1.000 |
| | BLX(0.5) | 2.265e-15 | 0.000 | 3.967e+01 | 0.000 | -4.179e+00 | 1.000 |
| | SBX(2) | -1.642e-12 | 0.000 | 3.948e+01 | 0.000 | -3.222e-01 | 1.000 |
| | SBX(5) | -4.871e-12 | 0.000 | 3.929e+01 | 0.000 | -3.678e+00 | 1.000 |
| | Logical | -3.668e-13 | 0.000 | 2.870e+01 | 0.000 | 3.957e-01 | 1.000 |
| | UNDX | -2.910e-05 | 0.000 | 1.888e+01 | 0.000 | -9.730e-01 | 1.000 |
| Logical | CIXL2 | 3.689e-13 | 0.000 | 1.098e+01 | 0.000 | 2.092e+00 | 0.000 |
| | BLX(0.3) | 3.692e-13 | 0.000 | 1.097e+01 | 0.000 | -2.191e+00 | 1.000 |
| | BLX(0.5) | 3.690e-13 | 0.000 | 1.098e+01 | 0.000 | -4.575e+00 | 1.000 |
| | SBX(2) | -1.275e-12 | 0.000 | 1.078e+01 | 0.000 | -7.179e-01 | 1.000 |
| | SBX(5) | -4.504e-12 | 0.000 | 1.059e+01 | 0.000 | -4.073e+00 | 1.000 |
| | Fuzzy | 3.668e-13 | 0.000 | -2.870e+01 | 0.000 | -3.957e-01 | 1.000 |
| | UNDX | -2.910e-05 | 0.000 | -9.812e+00 | 0.000 | -1.369e+00 | 0.000 |
| UNDX | CIXL2 | 2.910e-05 | 0.000 | 2.080e+01 | 0.000 | 3.460e+00 | 0.000 |
| | BLX(0.3) | 2.910e-05 | 0.000 | 2.078e+01 | 0.000 | -8.225e-01 | 1.000 |
| | BLX(0.5) | 2.910e-05 | 0.000 | 2.079e+01 | 0.000 | -3.206e+00 | 1.000 |
| | SBX(2) | 2.910e-05 | 0.000 | 2.060e+01 | 0.000 | 6.508e-01 | 1.000 |
| | SBX(5) | 2.910e-05 | 0.000 | 2.041e+01 | 0.000 | -2.705e+00 | 1.000 |
| | Fuzzy | 2.910e-05 | 0.000 | -1.888e+01 | 0.000 | 9.730e-01 | 1.000 |
| | Logical | 2.910e-05 | 0.000 | 9.812e+00 | 0.000 | 1.369e+00 | 0.000 |

| **Function** | **Ranking** |
|---|---|
| $f_{Sph}$ | $\mu_{UNDX} > \mu_{SBX(5)} > \mu_{SBX(2)} > \mu_{Logical} > \mu_{Ext.F.} > \mu_{CIXL2} \geq \mu_{BLX(0.5)} \geq \mu_{BLX(0.3)}$ |
| $f_{SchDS}$ | $\mu_{Ext.F.} > \mu_{UNDX} > \mu_{Logical} > \mu_{SBX(5)} \geq \mu_{SBX(2)} > \mu_{BLX(0.3)} \geq \mu_{BLX(0.5)} > \mu_{CIXL2}$ |
| $f_{Ros}$ | $\mu_{BLX(0.5)} \geq \mu_{SBX(5)} \geq \mu_{BLX(0.3)} \geq \mu_{UNDX} \geq \mu_{SBX(2)} \geq \mu_{Ext.F.} \geq \mu_{Logical} > \mu_{CIXL2}$ |

Table 14: Results of the multiple comparison tests for $f_{Sph}$, $f_{SchDS}$ y $f_{Ros}$ functions and the ranking established by the test regarding the crossover operator.

| Crossover | | $f_{Ras}$ | | $f_{Sch}$ | | $f_{Ack}$ | |
|---|---|---|---|---|---|---|---|
| **I** | **J** | $\mu_I - \mu_J$ | $\alpha^*$ | $\mu_I - \mu_J$ | $\alpha^*$ | $\mu_I - \mu_J$ | $\alpha^*$ |
| CIXL2 | BLX(0.3) | 7.296e-01 | 0.923 | 2.715e+02 | 0.000 | -2.830e-08 | 0.000 |
| | BLX(0.5) | -9.950e-02 | 1.000 | 2.210e+02 | 0.010 | -5.090e-08 | 0.000 |
| | SBX(2) | -1.552e+01 | 0.000 | -8.287e+02 | 0.000 | -5.322e-06 | 0.000 |
| | SBX(5) | -1.128e+01 | 0.000 | -4.631e+02 | 0.000 | -9.649e-06 | 0.000 |
| | Fuzzy | -1.953e+01 | 0.000 | -2.408e+03 | 0.000 | -1.659e-07 | 0.000 |
| | Logical | -6.033e+01 | 0.000 | -1.988e+03 | 0.000 | -2.517e-06 | 0.000 |
| | UNDX | -1.078e+02 | 0.000 | -7.409e+03 | 0.000 | -3.550e-02 | 0.000 |
| BLX(0.3) | CIXL2 | -7.296e-01 | 0.923 | -2.715e+02 | 0.000 | 2.830e-08 | 0.000 |
| | BLX(0.5) | -8.291e-01 | 0.713 | -5.050e+01 | 1.000 | -2.261e-08 | 0.000 |
| | SBX(2) | -1.625e+01 | 0.000 | -1.100e+03 | 0.000 | -5.293e-06 | 0.000 |
| | SBX(5) | -1.201e+01 | 0.000 | -7.346e+02 | 0.000 | -9.620e-06 | 0.000 |
| | Fuzzy | -2.026e+01 | 0.000 | -2.680e+03 | 0.000 | -1.376e-07 | 0.000 |
| | Logical | -6.106e+01 | 0.000 | -2.260e+03 | 0.000 | -2.488e-06 | 0.000 |
| | UNDX | -1.085e+02 | 0.000 | -7.680e+03 | 0.000 | -3.550e-02 | 0.000 |
| BLX(0.5) | CIXL2 | 9.950e-02 | 1.000 | -2.210e+02 | 0.010 | 5.090e-08 | 0.000 |
| | BLX(0.3) | 8.291e-01 | 0.713 | 5.050e+01 | 1.000 | 2.261e-08 | 0.000 |
| | SBX(2) | -1.542e+01 | 0.000 | -1.050e+03 | 0.000 | -5.271e-06 | 0.000 |
| | SBX(5) | -1.118e+01 | 0.000 | -6.841e+02 | 0.000 | -9.598e-06 | 0.000 |
| | Fuzzy | -1.943e+01 | 0.000 | -2.629e+03 | 0.000 | -1.150e-07 | 0.000 |
| | Logical | -6.023e+01 | 0.000 | -2.209e+03 | 0.000 | -2.466e-06 | 0.000 |
| | UNDX | -1.077e+02 | 0.000 | -7.630e+03 | 0.000 | -3.550e-02 | 0.000 |
| SBX(2) | CIXL2 | 1.552e+01 | 0.000 | 8.287e+02 | 0.000 | 5.322e-06 | 0.000 |
| | BLX(0.3) | 1.625e+01 | 0.000 | 1.100e+03 | 0.000 | 5.293e-06 | 0.000 |
| | BLX(0.5) | 1.542e+01 | 0.000 | 1.050e+03 | 0.000 | 5.271e-06 | 0.000 |
| | SBX(5) | 4.245e+00 | 0.005 | 3.655e+02 | 0.006 | -4.327e-06 | 0.000 |
| | Fuzzy | -4.013e+00 | 0.042 | -1.579e+03 | 0.000 | 5.156e-06 | 0.000 |
| | Logical | -4.481e+01 | 0.000 | -1.159e+03 | 0.000 | 2.805e-06 | 0.000 |
| | UNDX | -9.227e+01 | 0.000 | -6.580e+03 | 0.000 | -3.550e-02 | 0.000 |
| SBX(5) | CIXL2 | 1.128e+01 | 0.000 | 4.631e+02 | 0.000 | 9.649e-06 | 0.000 |
| | BLX(0.3) | 1.201e+01 | 0.000 | 7.346e+02 | 0.000 | 9.620e-06 | 0.000 |
| | BLX(0.5) | 1.118e+01 | 0.000 | 6.841e+02 | 0.000 | 9.598e-06 | 0.000 |
| | SBX(2) | -4.245e+00 | 0.005 | -3.655e+02 | 0.006 | 4.327e-06 | 0.000 |
| | Fuzzy | -8.258e+00 | 0.000 | -1.945e+03 | 0.000 | 9.483e-06 | 0.000 |
| | Logical | -4.905e+01 | 0.000 | -1.525e+03 | 0.000 | 7.132e-06 | 0.000 |
| | UNDX | -9.651e+01 | 0.000 | -6.946e+03 | 0.000 | -3.550e-02 | 0.000 |
| Fuzzy | CIXL2 | 1.953e+01 | 0.000 | 2.408e+03 | 0.000 | 1.659e-07 | 0.000 |
| | BLX(0.3) | 2.026e+01 | 0.000 | 2.680e+03 | 0.000 | 1.376e-07 | 0.000 |
| | BLX(0.5) | 1.943e+01 | 0.000 | 2.629e+03 | 0.000 | 1.150e-07 | 0.000 |
| | SBX(2) | 4.013e+00 | 0.042 | 1.579e+03 | 0.000 | -5.156e-06 | 0.000 |
| | SBX(5) | 8.258e+00 | 0.000 | 1.945e+03 | 0.000 | -9.483e-06 | 0.000 |
| | Logical | -4.079e+01 | 0.000 | 4.199e+02 | 0.000 | -2.351e-06 | 0.000 |
| | UNDX | -8.826e+01 | 0.000 | -5.001e+03 | 0.000 | -3.550e-02 | 0.000 |
| Logical | CIXL2 | 6.033e+01 | 0.000 | 1.988e+03 | 0.000 | 2.517e-06 | 0.000 |
| | BLX(0.3) | 6.106e+01 | 0.000 | 2.260e+03 | 0.000 | 2.488e-06 | 0.000 |
| | BLX(0.5) | 6.023e+01 | 0.000 | 2.209e+03 | 0.000 | 2.466e-06 | 0.000 |
| | SBX(2) | 4.481e+01 | 0.000 | 1.159e+03 | 0.000 | -2.805e-06 | 0.000 |
| | SBX(5) | 4.905e+01 | 0.000 | 1.525e+03 | 0.000 | -7.132e-06 | 0.000 |
| | Fuzzy | 4.079e+01 | 0.000 | -4.199e+02 | 0.000 | 2.351e-06 | 0.000 |
| | UNDX | -4.746e+01 | 0.000 | -5.421e+03 | 0.000 | -3.550e-02 | 0.000 |
| UNDX | CIXL2 | 1.078e+02 | 0.000 | 7.409e+03 | 0.000 | 3.550e-02 | 0.000 |
| | BLX(0.3) | 1.085e+02 | 0.000 | 7.680e+03 | 0.000 | 3.550e-02 | 0.000 |
| | BLX(0.5) | 1.077e+02 | 0.000 | 7.630e+03 | 0.000 | 3.550e-02 | 0.000 |
| | SBX(2) | 9.227e+01 | 0.000 | 6.580e+03 | 0.000 | 3.550e-02 | 0.000 |
| | SBX(5) | 9.651e+01 | 0.000 | 6.946e+03 | 0.000 | 3.550e-02 | 0.000 |
| | Fuzzy | 8.826e+01 | 0.000 | 5.001e+03 | 0.000 | 3.550e-02 | 0.000 |
| | Logical | 4.746e+01 | 0.000 | 5.421e+03 | 0.000 | 3.550e-02 | 0.000 |

| **Function** | **Ranking** |
|---|---|
| $f_{Ras}$ | $\mu_{UNDX} > \mu_{Logical} > \mu_{Ext.F.} > \mu_{SBX(2)} > \mu_{SBX(5)} > \mu_{BLX(0.5)} \geq \mu_{CIXL2} \geq \mu_{BLX(0.3)}$ |
| $f_{Sch}$ | $\mu_{UNDX} > \mu_{Ext.F.} > \mu_{Logical} > \mu_{SBX(2)} > \mu_{SBX(5)} > \mu_{CIXL2} > \mu_{BLX(0.5)} \geq \mu_{BLX(0.3)}$ |
| $f_{Ack}$ | $\mu_{UNDX} > \mu_{SBX(5)} > \mu_{SBX(2)} > \mu_{Logical} > \mu_{Ext.F.} > \mu_{BLX(0.5)} > \mu_{BLX(0.3)} > \mu_{CIXL2}$ |

Table 15: Results of the multiple comparison tests for $f_{Ras}$, $f_{Sch}$ and $f_{Ack}$ functions and the ranking established by the test regarding the crossover operator.

| Crossover | | $\mathbf{f_{Gri}}$ | | $\mathbf{f_{Fle}}$ | | $\mathbf{f_{Lan}}$ | |
|---|---|---|---|---|---|---|---|
| **I** | **J** | $\mu_\mathbf{I} - \mu_\mathbf{J}$ | $\alpha^*$ | $\mu_\mathbf{I} - \mu_\mathbf{J}$ | $\alpha^*$ | $\mu_\mathbf{I} - \mu_\mathbf{J}$ | $\alpha^*$ |
| CIXL2 | BLX(0.3) | -3.224e-02 | 0.021 | -4.779e+02 | 1.000 | 9.384e-02 | 0.091 |
| | BLX(0.5) | -2.235e-02 | 0.012 | -2.789e+03 | 1.000 | 1.392e-01 | 0.007 |
| | SBX(2) | -6.710e-03 | 0.973 | -1.740e+04 | 0.034 | -1.253e-02 | 1.000 |
| | SBX(5) | -1.603e-02 | 0.167 | -1.810e+04 | 0.022 | -1.982e-02 | 1.000 |
| | Fuzzy | 1.394e-02 | 0.000 | -1.686e+03 | 1.000 | -1.000e-01 | 0.000 |
| | Logical | 9.173e-03 | 0.057 | -1.196e+04 | 0.709 | -2.064e-01 | 0.000 |
| | UNDX | -6.312e-02 | 0.000 | -1.947e+04 | 0.009 | 6.557e-03 | 1.000 |
| BLX(0.3) | CIXL2 | 3.224e-02 | 0.021 | 4.779e+02 | 1.000 | -9.384e-02 | 0.091 |
| | BLX(0.5) | 9.893e-03 | 1.000 | -2.311e+03 | 1.000 | 4.540e-02 | 1.000 |
| | SBX(2) | 2.553e-02 | 0.188 | -1.693e+04 | 0.046 | -1.064e-01 | 0.046 |
| | SBX(5) | 1.621e-02 | 0.952 | -1.763e+04 | 0.029 | -1.137e-01 | 0.013 |
| | Fuzzy | 4.618e-02 | 0.000 | -1.208e+03 | 1.000 | -1.938e-01 | 0.000 |
| | Logical | 4.142e-02 | 0.001 | -1.148e+04 | 0.888 | -3.003e-01 | 0.000 |
| | UNDX | -3.088e-02 | 0.252 | -1.899e+04 | 0.012 | -8.728e-02 | 0.151 |
| BLX(0.5) | CIXL2 | 2.235e-02 | 0.012 | 2.789e+03 | 1.000 | -1.392e-01 | 0.007 |
| | BLX(0.3) | -9.893e-03 | 1.000 | 2.311e+03 | 1.000 | -4.540e-02 | 1.000 |
| | SBX(2) | 1.564e-02 | 0.361 | -1.461e+04 | 0.179 | -1.518e-01 | 0.004 |
| | SBX(5) | 6.320e-03 | 1.000 | -1.531e+04 | 0.121 | -1.591e-01 | 0.001 |
| | Fuzzy | 3.629e-02 | 0.000 | 1.104e+03 | 1.000 | -2.392e-01 | 0.000 |
| | Logical | 3.152e-02 | 0.000 | -9.169e+03 | 1.000 | -3.457e-01 | 0.000 |
| | UNDX | -4.077e-02 | 0.003 | -1.668e+04 | 0.054 | -1.327e-01 | 0.012 |
| SBX(2) | CIXL2 | 6.710e-03 | 0.973 | 1.740e+04 | 0.034 | 1.253e-02 | 1.000 |
| | BLX(0.3) | -2.553e-02 | 0.188 | 1.693e+04 | 0.046 | 1.064e-01 | 0.046 |
| | BLX(0.5) | -1.564e-02 | 0.361 | 1.461e+04 | 0.179 | 1.518e-01 | 0.004 |
| | SBX(5) | -9.320e-03 | 0.980 | -7.002e+02 | 1.000 | -7.285e-03 | 1.000 |
| | Fuzzy | 2.065e-02 | 0.000 | 1.572e+04 | 0.095 | -8.747e-02 | 0.008 |
| | Logical | 1.588e-02 | 0.003 | 5.446e+03 | 1.000 | -1.939e-01 | 0.000 |
| | UNDX | -5.641e-02 | 0.000 | -2.061e+03 | 1.000 | 1.909e-02 | 1.000 |
| SBX(5) | CIXL2 | 1.603e-02 | 0.167 | 1.810e+04 | 0.022 | 1.982e-02 | 1.000 |
| | BLX(0.3) | -1.621e-02 | 0.952 | 1.763e+04 | 0.029 | 1.137e-01 | 0.013 |
| | BLX(0.5) | -6.320e-03 | 1.000 | 1.531e+04 | 0.121 | 1.591e-01 | 0.001 |
| | SBX(2) | 9.320e-03 | 0.980 | 7.002e+02 | 1.000 | 7.285e-03 | 1.000 |
| | Fuzzy | 2.997e-02 | 0.000 | 1.642e+04 | 0.063 | -8.018e-02 | 0.004 |
| | Logical | 2.520e-02 | 0.001 | 6.146e+03 | 1.000 | -1.866e-01 | 0.000 |
| | UNDX | -4.709e-02 | 0.000 | -1.361e+03 | 1.000 | 2.637e-02 | 1.000 |
| Fuzzy | CIXL2 | -1.394e-02 | 0.000 | 1.686e+03 | 1.000 | 1.000e-01 | 0.000 |
| | BLX(0.3) | -4.618e-02 | 0.000 | 1.208e+03 | 1.000 | 1.938e-01 | 0.000 |
| | BLX(0.5) | -3.629e-02 | 0.000 | -1.104e+03 | 1.000 | 2.392e-01 | 0.000 |
| | SBX(2) | -2.065e-02 | 0.000 | -1.572e+04 | 0.095 | 8.747e-02 | 0.008 |
| | SBX(5) | -2.997e-02 | 0.000 | -1.642e+04 | 0.063 | 8.018e-02 | 0.004 |
| | Logical | -4.763e-03 | 0.025 | -1.027e+04 | 1.000 | -1.064e-01 | 0.000 |
| | UNDX | -7.706e-02 | 0.000 | -1.778e+04 | 0.027 | 1.066e-01 | 0.000 |
| Logical | CIXL2 | -9.173e-03 | 0.057 | 1.196e+04 | 0.709 | 2.064e-01 | 0.000 |
| | BLX(0.3) | -4.142e-02 | 0.001 | 1.148e+04 | 0.888 | 3.003e-01 | 0.000 |
| | BLX(0.5) | -3.152e-02 | 0.000 | 9.169e+03 | 1.000 | 3.457e-01 | 0.000 |
| | SBX(2) | -1.588e-02 | 0.003 | -5.446e+03 | 1.000 | 1.939e-01 | 0.000 |
| | SBX(5) | -2.520e-02 | 0.001 | -6.146e+03 | 1.000 | 1.866e-01 | 0.000 |
| | Fuzzy | 4.763e-03 | 0.025 | 1.027e+04 | 1.000 | 1.064e-01 | 0.000 |
| | UNDX | -7.229e-02 | 0.000 | -7.507e+03 | 1.000 | 2.130e-01 | 0.000 |
| UNDX | CIXL2 | 6.312e-02 | 0.000 | 1.947e+04 | 0.009 | -6.557e-03 | 1.000 |
| | BLX(0.3) | 3.088e-02 | 0.252 | 1.899e+04 | 0.012 | 8.728e-02 | 0.151 |
| | BLX(0.5) | 4.077e-02 | 0.003 | 1.668e+04 | 0.054 | 1.327e-01 | 0.012 |
| | SBX(2) | 5.641e-02 | 0.000 | 2.061e+03 | 1.000 | -1.909e-02 | 1.000 |
| | SBX(5) | 4.709e-02 | 0.000 | 1.361e+03 | 1.000 | -2.637e-02 | 1.000 |
| | Fuzzy | 7.706e-02 | 0.000 | 1.778e+04 | 0.027 | -1.066e-01 | 0.000 |
| | Logical | 7.229e-02 | 0.000 | 7.507e+03 | 1.000 | -2.130e-01 | 0.000 |
| **Function** | **Ranking** | | | | | | |
| $f_{Gri}$ | $\mu_{UNDX} \geq \mu_{BLX(0.3)} \geq \mu_{BLX(0.5)} \geq \mu_{SBX(5)} \geq \mu_{SBX(2)} \geq \mu_{CIXL2} \geq \mu_{Logical} > \mu_{Ext.F.}$ | | | | | | |
| $f_{Fle}$ | $\mu_{UNDX} \geq \mu_{SBX(5)} \geq \mu_{SBX(2)} \geq \mu_{Logical} \geq \mu_{BLX(0.5)} \geq \mu_{Ext.F.} \geq \mu_{BLX(0.3)} \geq \mu_{CIXL2}$ | | | | | | |
| $f_{Lan}$ | $\mu_{Logical} > \mu_{Ext.F.} > \mu_{SBX(5)} \geq \mu_{SBX(2)} \geq \mu_{CIXL2} \geq \mu_{UNDX} \geq \mu_{BLX(0.3)} \geq \mu_{BLX(0.5)}$ | | | | | | |

Table 16: Results of the multiple comparison tests for $f_{Gri}$, $f_{Fle}$ and $f_{Lan}$ functions and the ranking established by the test regarding the crossover operator.

| I | J | $\mu_I - \mu_J$ | $\alpha^*$ | $\mu_I - \mu_J$ | $\alpha^*$ | $\mu_I - \mu_J$ | $\alpha^*$ | $\mu_I - \mu_J$ | $\alpha^*$ |
|---|---|---|---|---|---|---|---|---|---|
| | | $f_{SchDS}$ | | $f_{Ros}$ | | $f_{Ras}$ | | $f_{Sch}$ | |
| CIXL2 | $UMDA_c$ | -2.221e+01 | 0.000 | -2.928e+00 | 0.000 | -1.547e+02 | 0.000 | -1.089e+04 | 0.000 |
| | $EGNA_{BGe}$ | -2.076e-01 | 0.000 | -2.906e+00 | 0.000 | -1.533e+02 | 0.000 | -1.091e+04 | 0.000 |
| $UMDA_c$ | CIXL2 | 2.221e+01 | 0.000 | 2.928e+00 | 0.000 | 1.547e+02 | 0.000 | 1.089e+04 | 0.000 |
| | $EGNA_{BGe}$ | 2.200e+01 | 0.000 | 2.207e-02 | 0.856 | 1.360e+00 | 0.888 | -2.390e+01 | 0.677 |
| $EGNA_{BGe}$ | CIXL2 | 2.076e-01 | 0.000 | 2.906e+00 | 0.000 | 1.533e+02 | 0.000 | 1.091e+04 | 0.000 |
| | $UMDA_c$ | -2.200e+01 | 0.000 | -2.207e-02 | 0.856 | -1.360e+00 | 0.888 | 2.390e+01 | 0.677 |
| **Function** | | | | **Ranking** | | | | | |
| $f_{SchDS}$ | | | | $\mu_{UMDA_c} > \mu_{EGNA_{BGe}} > \mu_{CIXL2}$ | | | | | |
| $f_{Ros}$ | | | | $\mu_{UMDA_c} \geq \mu_{EGNA_{BGe}} > \mu_{CIXL2}$ | | | | | |
| $f_{Ras}$ | | | | $\mu_{UMDA_c} \geq \mu_{EGNA_{BGe}} > \mu_{CIXL2}$ | | | | | |
| $f_{Sch}$ | | | | $\mu_{EGNA_{BGe}} \geq \mu_{UMDA_c} > \mu_{CIXL2}$ | | | | | |
| | | $f_{Ack}$ | | $f_{Gri}$ | | $f_{Fle}$ | | $f_{Lan}$ | |
| CIXL2 | $UMDA_c$ | -1.101e-08 | 0.000 | 1.525e-02 | 0.000 | 9.803e+03 | 0.004 | -3.306e-02 | 0.176 |
| | $EGNA_{BGe}$ | -9.194e-09 | 0.000 | 1.525e-02 | 0.000 | 6.157e+03 | 0.150 | -3.306e-02 | 0.176 |
| $UMDA_c$ | CIXL2 | 1.101e-08 | 0.000 | -1.525e-02 | 0.000 | -9.803e+03 | 0.004 | 3.306e-02 | 0.176 |
| | $EGNA_{BGe}$ | 1.817e-09 | 0.175 | 1.266e-16 | 0.000 | -3.646e+03 | 0.049 | 1.33781e-11 | 0.325 |
| $EGNA_{BGe}$ | CIXL2 | 9.194e-09 | 0.000 | -1.525e-02 | 0.000 | -6.157e+03 | 0.150 | 3.306e-02 | 0.176 |
| | $UMDA_c$ | -1.817e-09 | 0.175 | -1.266e-16 | 0.000 | 3.646e+03 | 0.049 | -1.33781e-11 | 0.325 |
| **Function** | | | | **Ranking** | | | | | |
| $f_{Ack}$ | | | | $\mu_{UMDA_c} \geq \mu_{EGNA_{BGe}} > \mu_{CIXL2}$ | | | | | |
| $f_{Gri}$ | | | | $\mu_{CIXL2} > \mu_{UMDA_c} > \mu_{EGNA_{BGe}}$ | | | | | |
| $f_{Fle}$ | | | | $\mu_{CIXL2} \geq \mu_{EGNA_{BGe}} > \mu_{UMDA_c}$ | | | | | |
| $f_{Lan}$ | | | | $\mu_{UMDA_c} \geq \mu_{EGNA_{BGe}} \geq \mu_{CIXL2}$ | | | | | |

Table 17: Results for all the functions of the multiple comparison test and the ranking obtained depending on the evolutionary algorithm.

## Appendix B. Convergence Graphics

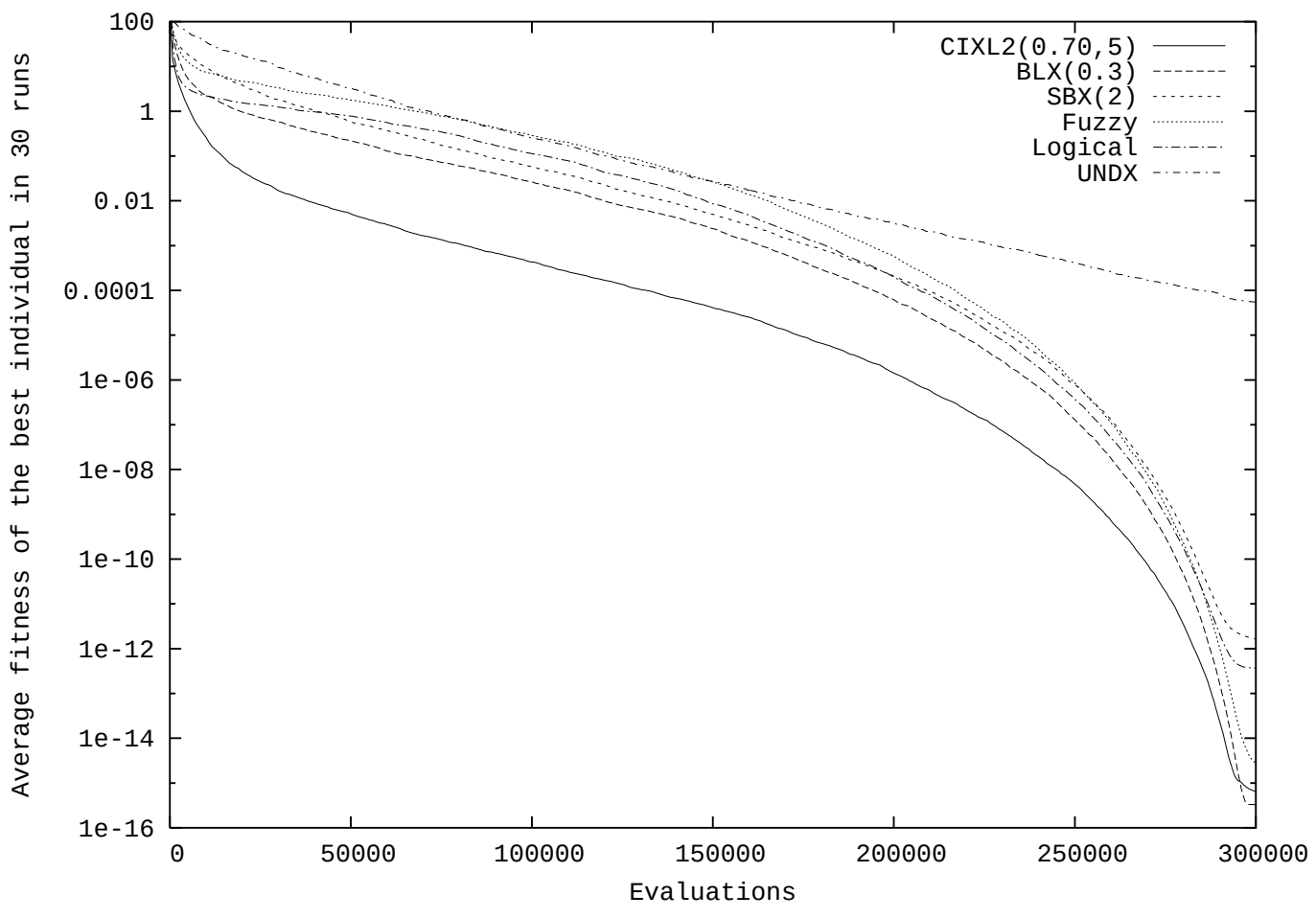

Figure 5: Evolution of the average fitness, in logarithmic scale, using different crossover operators for the function $f_{Sph}$.

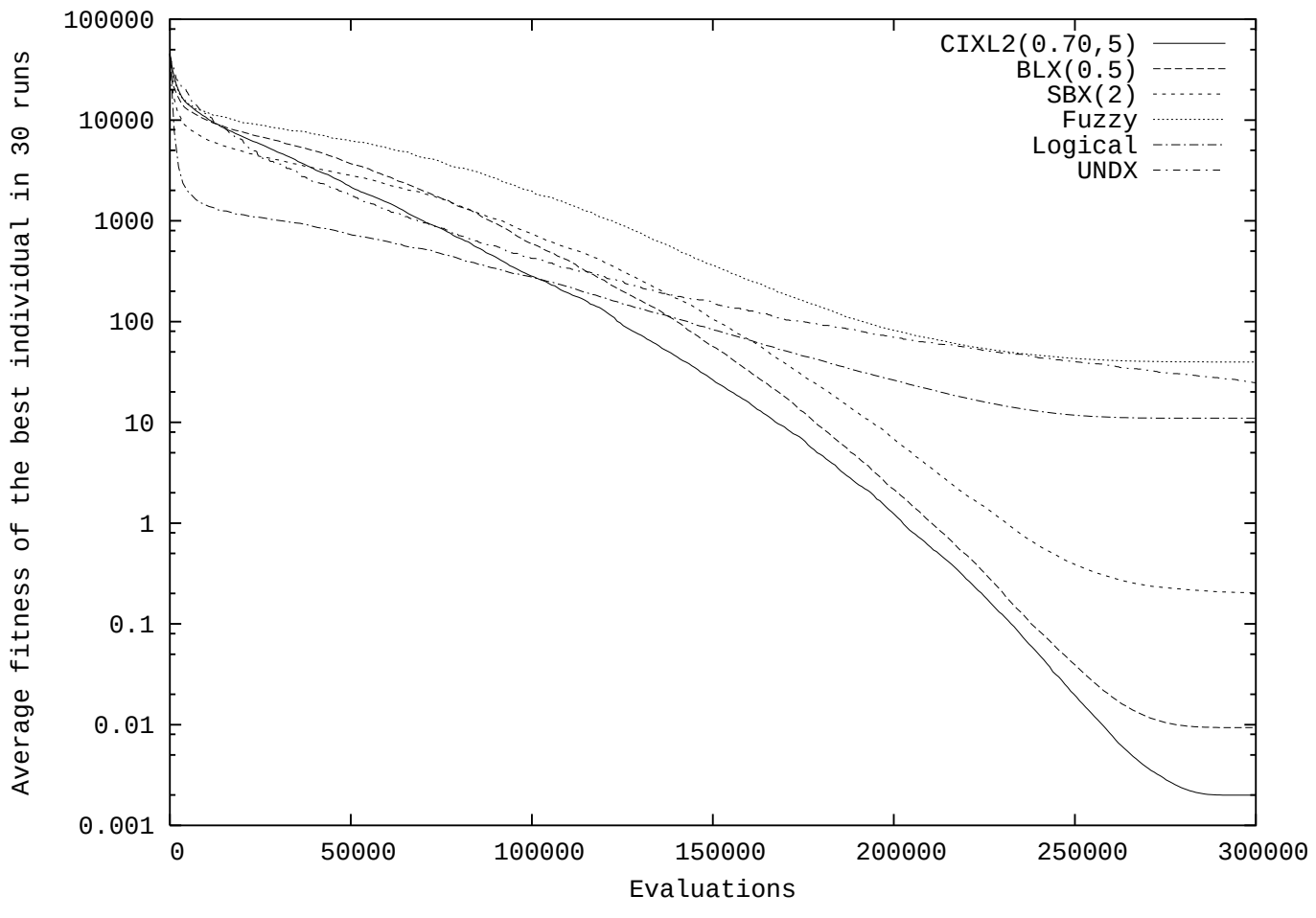

Figure 6: Evolution of the average fitness, in logarithmic scale, using different crossover operators for the function $f_{SchDS}$.

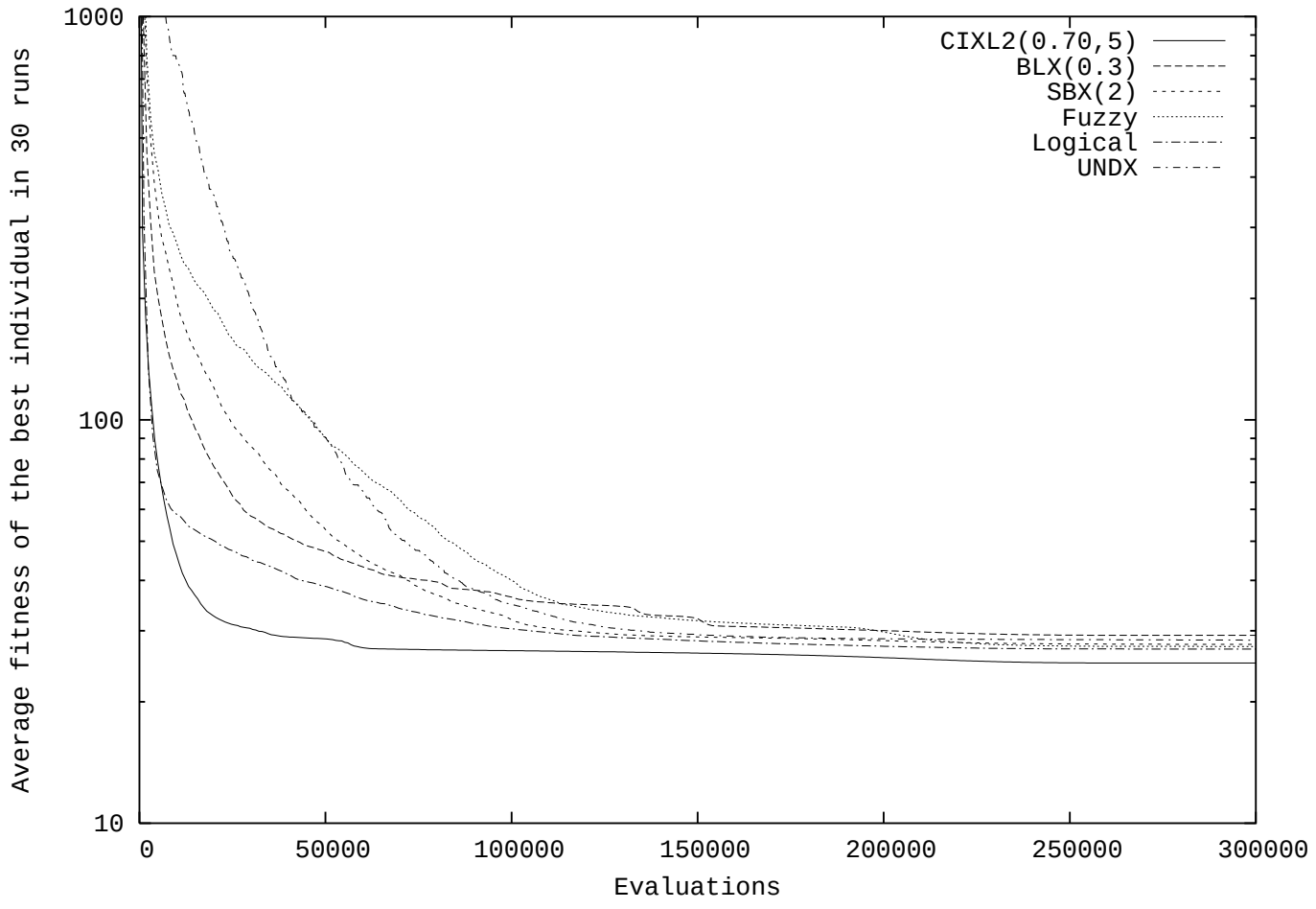

Figure 7: Evolution of the averaged fitness, in logarithmic scale, using different crossover operators for the function $f_{Ros}$.

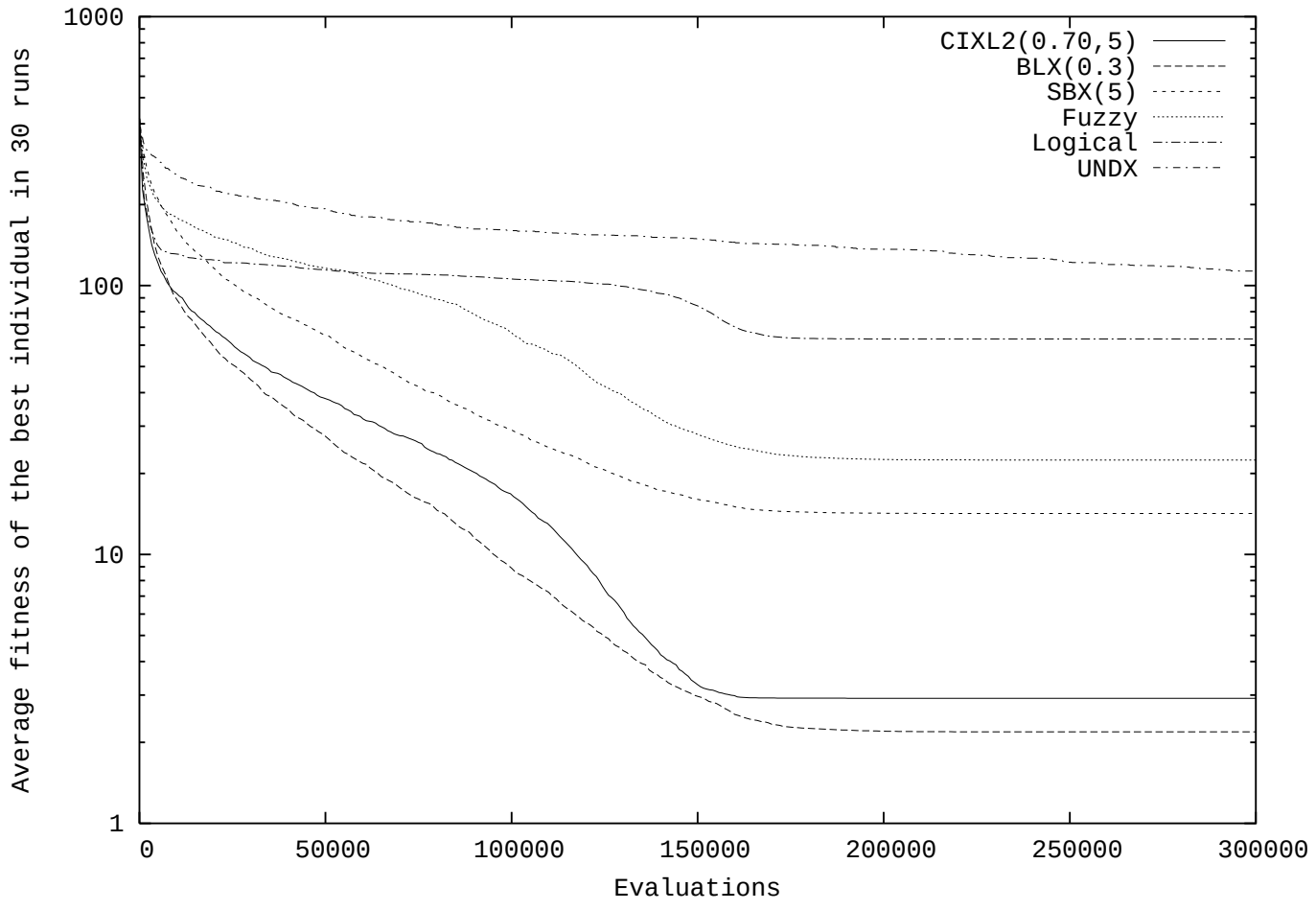

Figure 8: Evolution of the average fitness, in logarithmic scale, using different crossover operators for the function $f_{Ras}$.

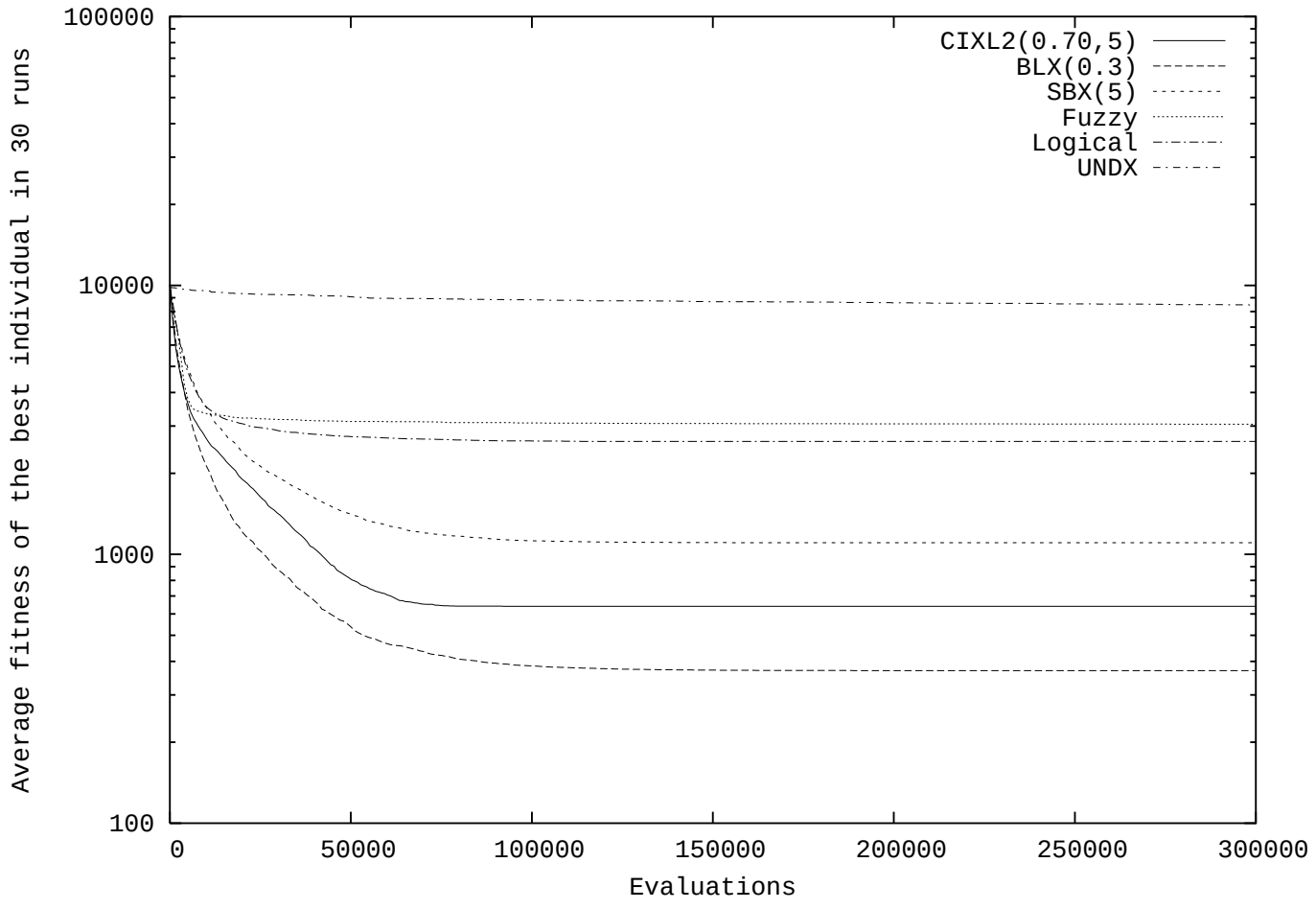

Figure 9: Evolution of the average fitness, in logarithmic scale, using different crossover operators for the function $f_{Sch}$.

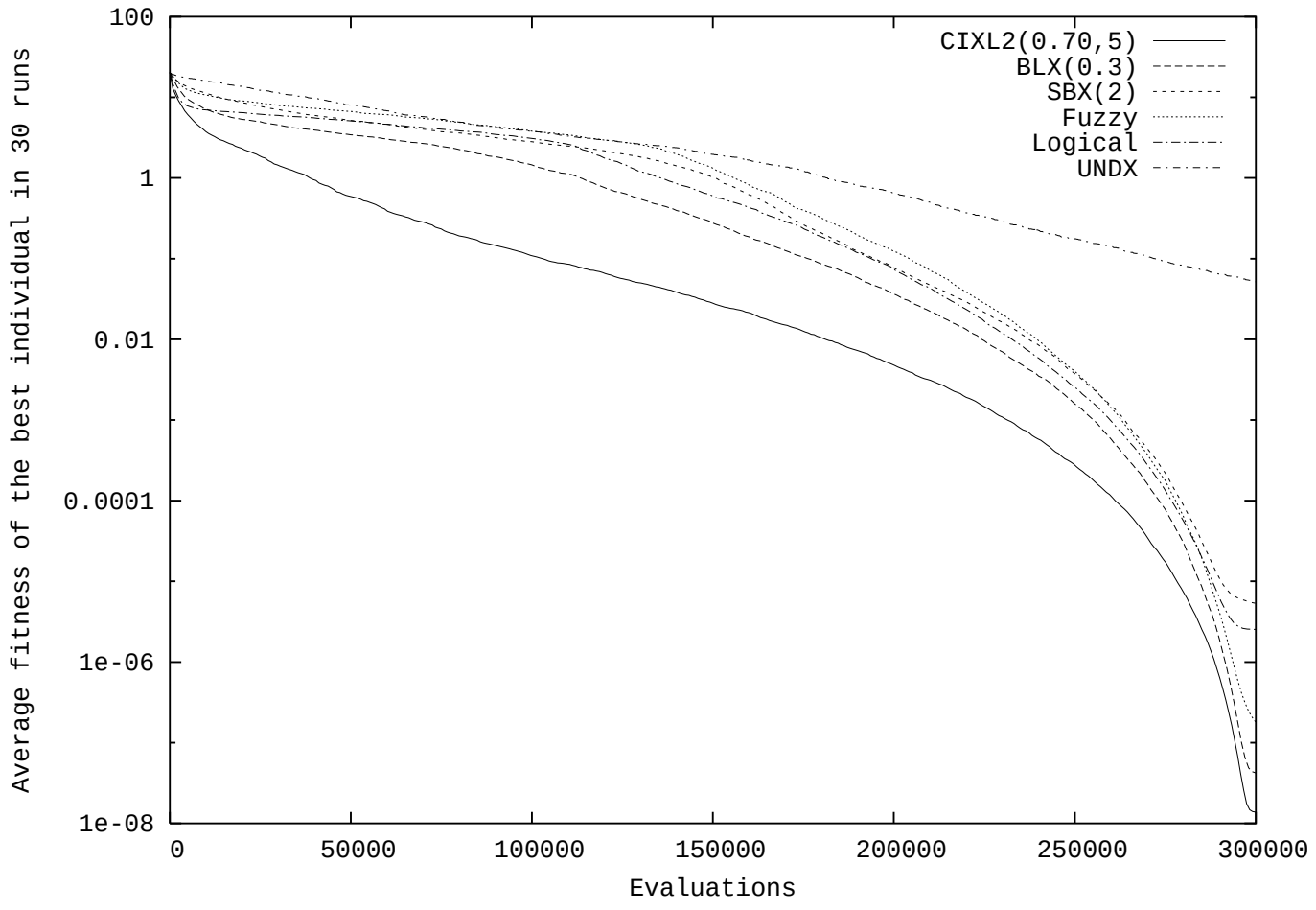

Figure 10: Evolution of the average fitness, in logarithmic scale, using different crossover operators for the function $f_{Ack}$.

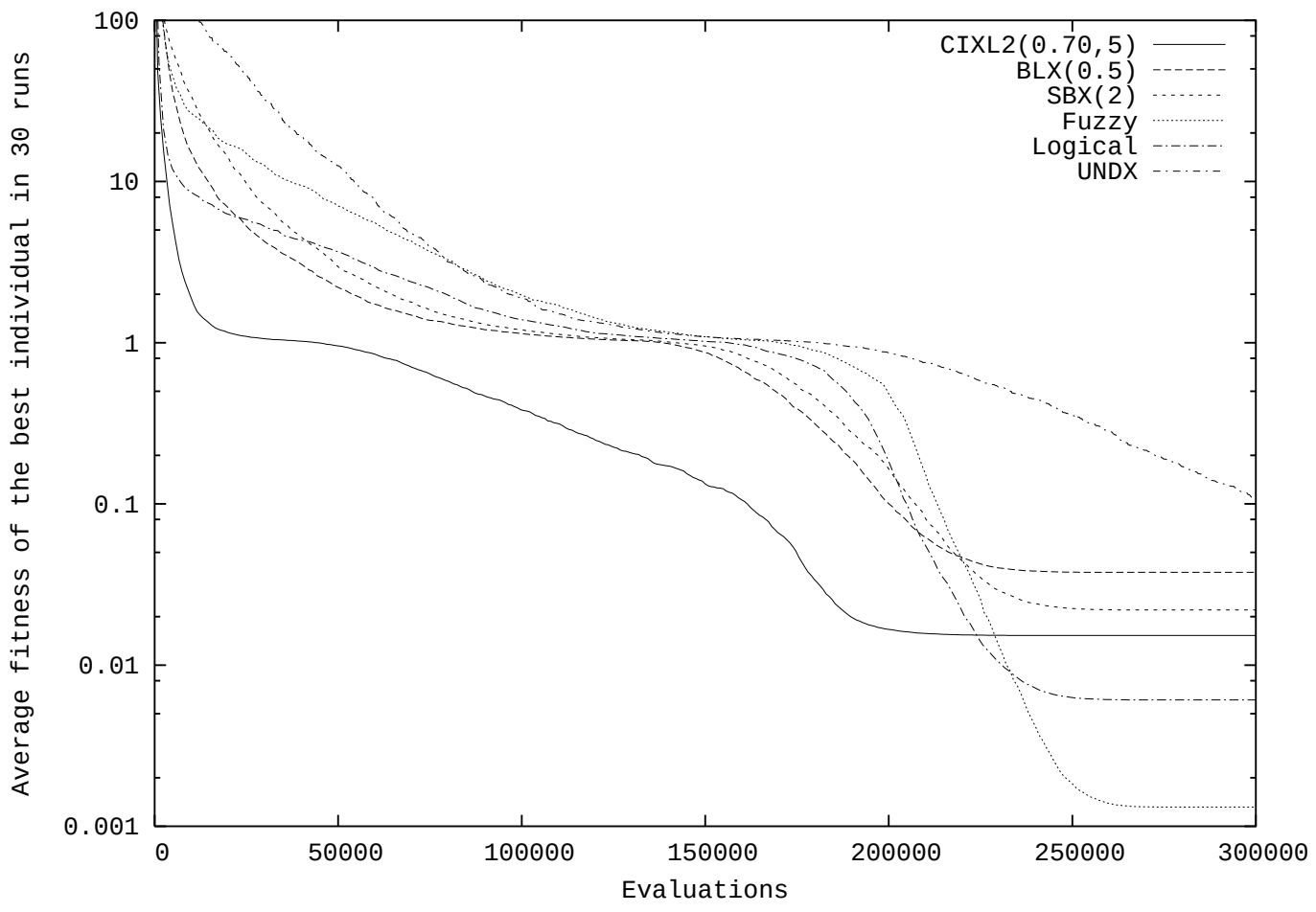

Figure 11: Evolution of the average fitness, in logarithmic scale, using different crossover operators for the function $f_{Gri}$.

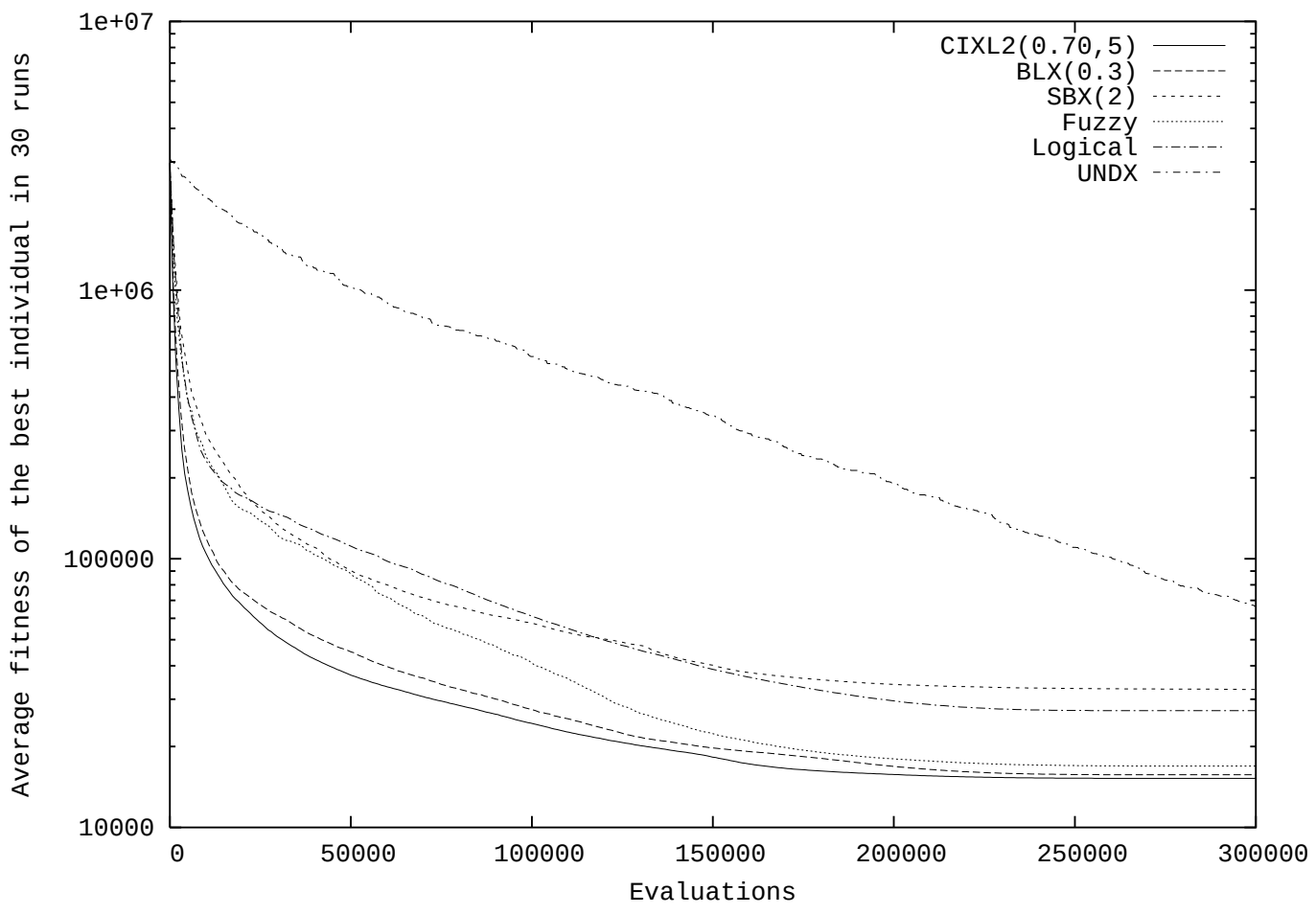

Figure 12: Evolution of the average, in logarithmic scale, using different crossover operators for the function $f_{Fle}$.

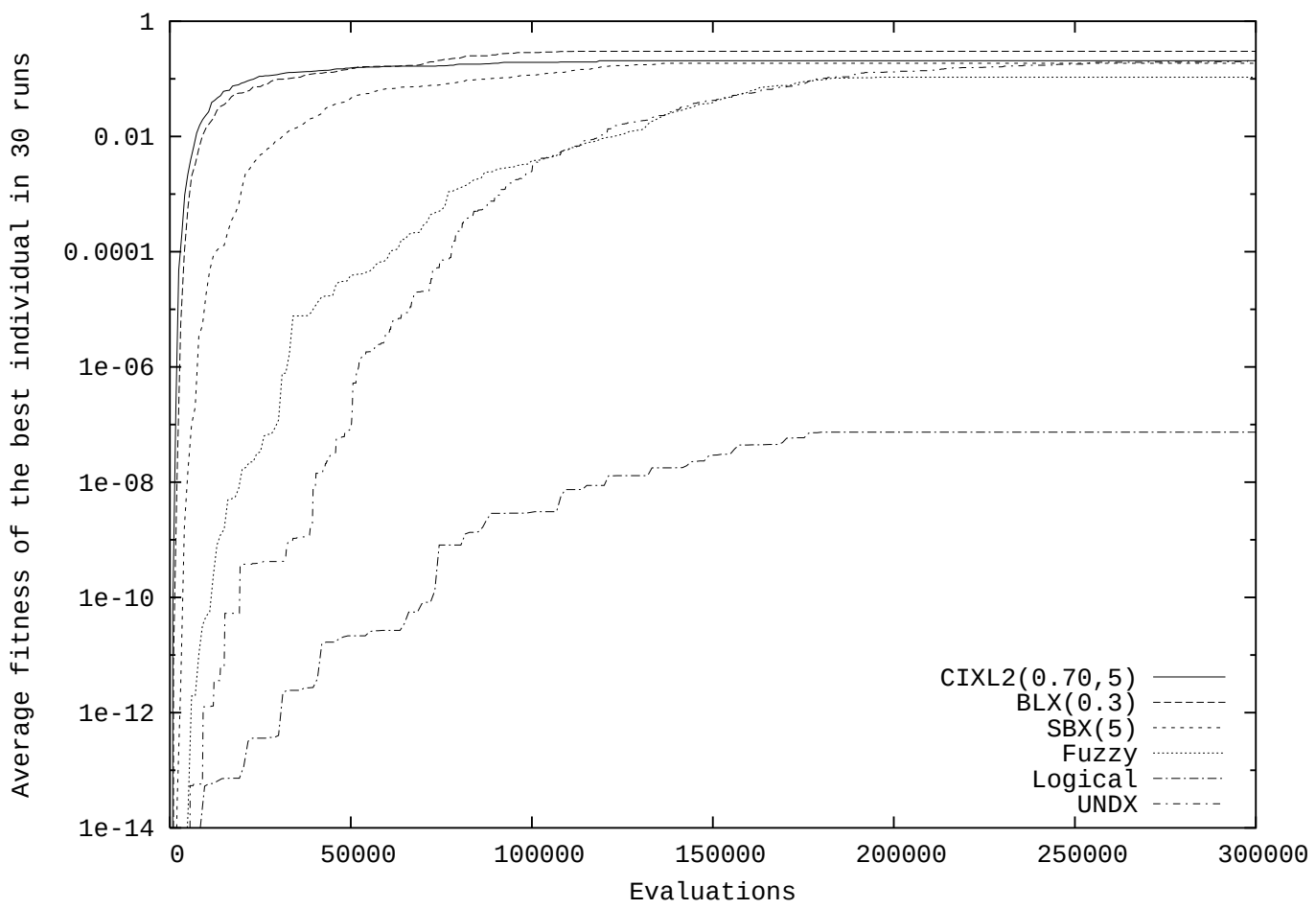

Figure 13: Evolution of the average fitness, in logarithmic scale, using different crossover operators for the function $f_{Lan}$.